\documentclass[review]{elsarticle}

\usepackage{lineno,hyperref}
\usepackage{amsmath, graphicx, multicol, multirow}
\usepackage[normalem]{ulem}
\useunder{\uline}{\ul}{}
\usepackage{rotating}
\modulolinenumbers[5]
\usepackage[table]{xcolor}
\journal{Medical Image Analysis}

\usepackage{numcompress}
\biboptions{authoryear}

\begin{document}
	
	\begin{frontmatter}
		
		\title{Micro-Net: A unified model for segmentation of various objects in microscopy images}
		
		
		\author[ICRaddress,DCSWarwick]{Shan E Ahmed Raza\corref{correspondingauthor}}
		\cortext[correspondingauthor]{Corresponding author}
		\ead{shan.raza@icr.ac.uk}
		\author[SLSWarwick]{Linda Cheung}
		\author[DCSWarwick]{Muhammad Shaban}
		\author[DCSWarwick]{Simon Graham}
		\author[MIWarwick]{David Epstein}
		\author[SLSWarwick]{Stella Pelengaris}
		\author[SLSWarwick]{Michael Khan}
		\author[DCSWarwick,DPUHCW,ATIUK]{Nasir M. Rajpoot\corref{correspondingauthor}}
		\ead{nasir.rajpoot@ieee.org}
		
		\address[ICRaddress]{Division of Molecular Pathology, The Institute of Cancer Research, UK.}
		\address[DCSWarwick]{Department of Computer Science, University of Warwick, UK.}
		\address[SLSWarwick]{School of Life Sciences, University of Warwick, UK.}
		\address[MIWarwick]{Department of Mathematics, University of Warwick, UK.}
		\address[DPUHCW]{Department of Pathology, University Hospitals Coventry and Warwickshire, UK.}
		\address[ATIUK]{The Alan Turing Institute, London, UK.}

		\begin{abstract}
			Object segmentation and structure localization are important steps in automated image analysis pipelines for microscopy images. We present a convolution neural network (CNN) based deep learning architecture for segmentation of objects in microscopy images. The proposed network can be used to segment cells, nuclei and glands in fluorescence microscopy and histology images after slight tuning of input parameters. The network trains at multiple resolutions of the input image, connects the intermediate layers for better localization and context and generates the output using multi-resolution deconvolution filters. The extra convolutional layers which bypass the max-pooling operation allow the network to train for variable input intensities and object size and make it robust to noisy data. We compare our results on publicly available data sets and show that the proposed network outperforms recent deep learning algorithms. 
		\end{abstract}
		
		
	\end{frontmatter}
	
	
	\section{Introduction}
	\label{sec:intro}
	In automated microscopic image analysis pipelines, segmentation of key structures such as tumours, glands and cells is an important step (\cite{awan2017glandular,yuan2012quantitative,qaiser2017tumor}). Recent advances in deep learning have helped to achieve accurate segmentation of these structures. A major strength of deep learning is that the same network architecture can be used to segment various structures across different modalities by retraining and slight tuning of the input parameters (\cite{FCN2016,Ronneberger2015}). 
	
\begin{figure}[tbh!]
		\centering
		\includegraphics[width=0.99\textwidth]{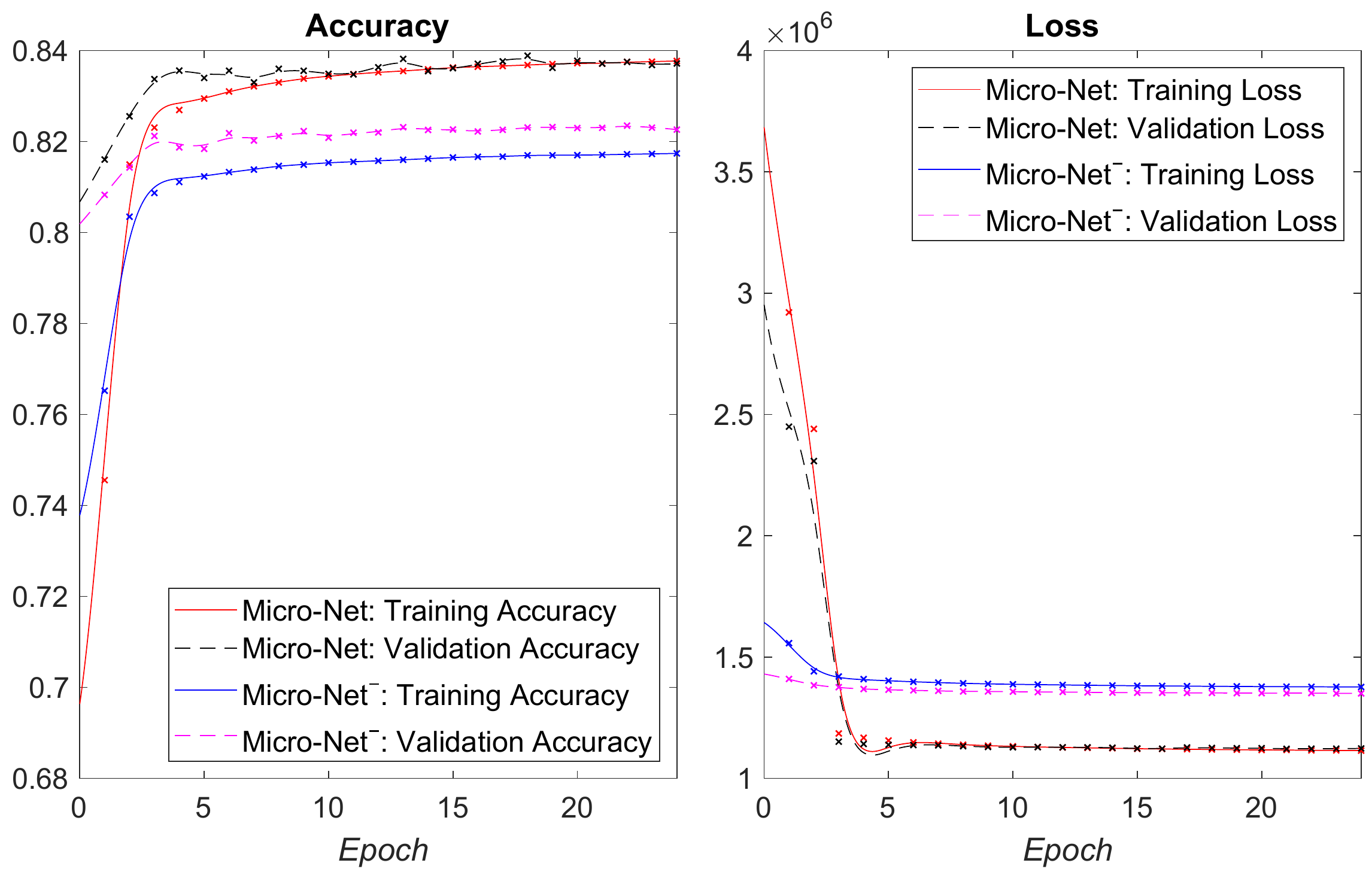}
		\caption{Solid lines represent training accuracy/loss for Micro-Net and Micro-Net\textsuperscript{--} while the dashed lines represent validation accuracy/loss calculated for 25 epochs on fluorescence imaging data for cell segmentation. Micro-Net\textsuperscript{--} was obtained by removing multi-resolution input and the bypass layers from Micro-Net architecture. The accuracy and loss curves clearly show the importance of the multi-resolution input and bypass layers.}
		\label{fig:MIMOvsSMIO}
	\end{figure}

\begin{figure}[tbh!]
		\centering
		\includegraphics[width=0.99\textwidth]{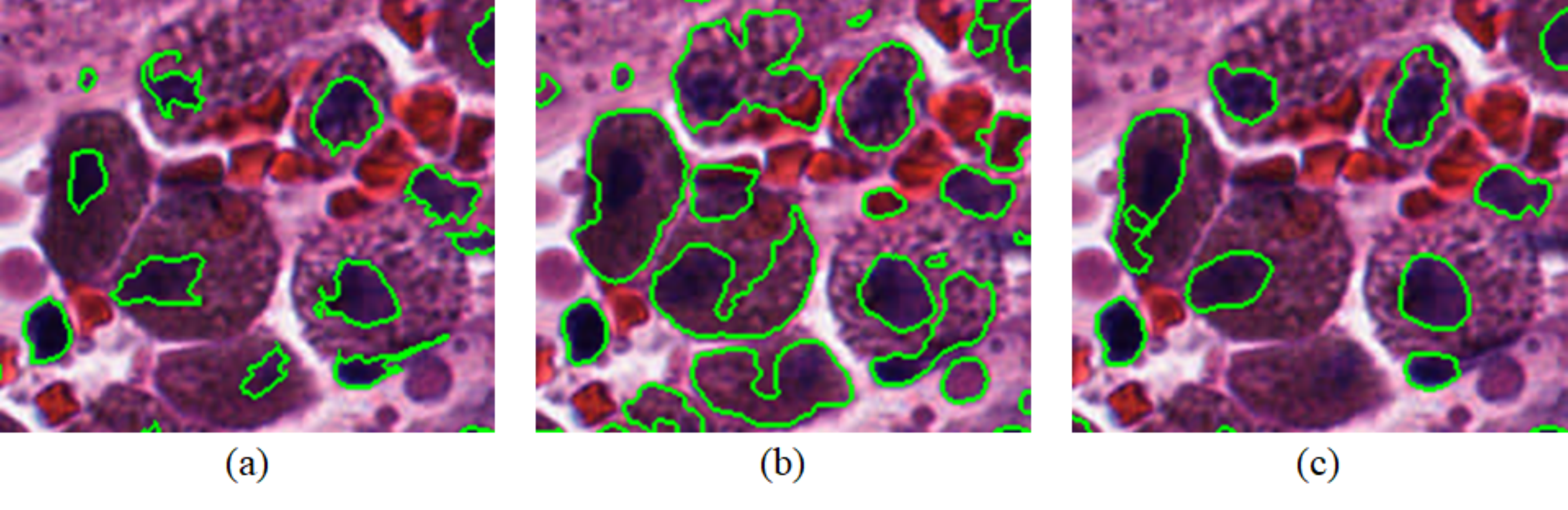}
		\caption{Nuclear segmentation on a sample H \& E image from lung outlined in green (a) ground truth, (b) U-Net (\cite{Ronneberger2015}) (c) proposed. U-Net clearly misses the boundary and is inclined towards strong contrast whereas the proposed method segments nuclei instead of strong contrast with the background. This is because the proposed approach learns the features at multiple input resolutions and learns for weaker boundaries.}
		\label{fig:Concept}
	\end{figure}

	In this paper, we propose a CNN with additional layers in the downsampling path, bypassing the max-pooling operation in order to learn the parameters for segmentation ignored during the max-pooling operation. By doing so, we retain contextual information, make the network interpret the output at multiple resolutions and train the model at multiple input image resolutions in the downsampling path to learn the model parameters for variable cell/nucleus/gland sizes and shapes in the presence of variable intensities and texture. There are two main features of the proposed architecture: (a) it learns image features at multiple input resolutions for better understanding of tissue components and (b) it bypasses the max-pooling operation through extra layers to retain information from weak features may be missed during max-pooling. This makes the network robust to noise and helps to learn the context at multiple resolutions. Figure \ref{fig:MIMOvsSMIO} \& \ref{fig:Concept} demonstrate the impact of these design changes. In Figure \ref{fig:MIMOvsSMIO}, solid lines represent \emph{training} accuracy/loss for Micro-Net and Micro-Net\textsuperscript{--}, whereas dashed lines represent \emph{validation} accuracy/loss for Micro-Net and Micro-Net\textsuperscript{--} during training. Accuracy is defined in terms of pixel-wise agreement with the ground truth and loss is defined in Section \ref{sec:lossfuntion}. In Micro-Net\textsuperscript{--}, we removed the multi-resolution input and the bypass layers while keeping the rest of the architecture the same. The improved accuracy and loss values demonstrate the importance of the proposed design changes. To emphasise this further, Figure \ref{fig:Concept}(a) shows an H \& E image where nuclei are outlined with green boundaries by an expert, Figure \ref{fig:Concept}(b) outlines the result of U-Net (\cite{Ronneberger2015}) and Figure \ref{fig:Concept}(c) the result of the proposed approach. It can be observed that U-Net failed to learn the features due to the presence of a dark cytoplasmic region and segmented most of the cellular region instead of just the nucleus, whereas the proposed approach learned the context at multiple resolutions and successfully located the nuclei despite high levels of noise. We discuss this in detail in Section \ref{sec:results}. 
		
	This paper is an extension of our previous work on cell (\cite{RazaISBI2017})\footnote {We will publish our fluorescence cell segmentation data set along with ground truth annotations subject to the publication of this manuscript at \href{http://go.warwick.ac.uk/tialab/data}{go.warwick.ac.uk/tialab/data}.} and gland segmentation (\cite{RazaMIUA2017}) with the following novel contributions:
	\begin{enumerate}
		\item A unified framework for segmentation of various types of objects (nuclei, cells, glands) in two different types of image modalities (histology and fluorescence microscopy).
		\item We discuss in detail the challenges faced for training a CNN for segmentation and present a solution to overcome those challenges.
		\item Detailed results to show the robustness of the method to high levels of noise and comparative evaluation with the state-of-the-art.
		\item We propose how the proposed network architecture can be modified/extended for different applications.
		\item In order to justify and clearly demonstrate the effect of additional layers, we present results for Micro-Net\textsuperscript{--} after removing the multi-resolution input and the bypass layers while keeping the rest of the architecture the same as Micro-Net.
		\item Addition of another data set to our analysis where we compare our results with those in the MICCAI 2017 computational precision medicine (CPM) challenge contest dataset for nuclear segmentation. 
		
	\end{enumerate}

	\subsection{Related Work}
	The existing literature on segmentation methods can be broadly classified into two main categories: handcrafted feature based approaches and deep learning based methods. Most of the existing handcrafted feature based approaches to cell/nuclear segmentation employ a combination of thresholding, filtering, morphological operations, region accumulation, marker controlled watershed (\cite{yang2006nuclei,Veta2013}), deformable model fitting (\cite{bergeest2012efficient}), graph cut (\cite{Dimopoulos15092014}) and feature classification (\cite{Li2015}). A detailed review of cell/nuclear segmentation methods was presented by \cite{Meijering2012} for images from various modalities. For gland segmentation, most of the early attempts used handcrafted features. \cite{Wu2005} identified initial seed regions based on large vacant lumen regions and expanded the seed to a surrounding chain of epithelial nuclei. \cite{Farjam2007} proposed segmentation by clustering texture features calculated using a variance filter. However, robust segmentation requires more domain knowledge and texture features calculated using just a variance filter might not provide enough information for the local structure of the tissue. \cite{naik2008automated} employed a Bayesian classifier to detect lumen regions and then refined using a level set to stop the curve, based on the likelihood of a nucleus. While this approach is reported to work well in benign cases, it can fail in malignant cases where the morphology of glands is quite complex. \cite{nguyen2012structure} grouped the nuclei, cytoplasm and lumen using colour space analysis and grew the lumen region with constraints to achieve segmentation. \cite{gunduz2010automatic} represented each tissue component as a circular disc and constructed a graph with nearby discs joined by an edge. They performed region growing on lumen discs that were constrained by lines joining the nuclear discs. \cite{nosrati2014local} and \cite{cohen2015memory} first classify tissue regions into different constituents and then employ a constrained level set algorithm to segment the glands. \cite{sirinukunwattana2015stochastic} identified epithelial superpixels and used epithelial regions as vertices of a polygon approximating the boundary of a gland. Most of the methods discussed above first distinguish tissue regions and then employ region growing or level sets to segment glandular regions. Recently, \cite{Li2017multi} proposed a slightly different approach where they first determine potential epithelial regions using lumen/background information and then identify connected epithelial cells to segment the glands using a multi-resolution cell orientation descriptor.
	
	In this paper, we focus on deep learning based approaches using convolutional neural networks (CNNs). These have recently received a wealth of attention, due to state-of-the-art performance in recent computer vision tasks, including segmentation (\cite{FCN2016,Ronneberger2015,Chen2017dcan,Song2016}). The fully convolutional network (FCN) for segmentation is considered to be a benchmark for segmentation tasks using deep learning (\cite{FCN2016}). The network performs pixel-wise classification to obtain the segmentation mask for a given input and consists of downsampling and upsampling paths. The downsampling path consists of convolution and max-pooling and the upsampling path consists of convolution and deconvolution (convolution transpose) layers. U-Net (\cite{Ronneberger2015}) is inspired by FCN but connects intermediate downsampling and upsampling paths to conserve the context information. Recently, \cite{sadanandan2017}  used the CellProfiler pipeline (\cite{cellprofiler2006}) as an automatic way of generating ground truth to train the network and employed a variation of fully convolutional network inspired by the improvements in U-Net (\cite{Ronneberger2015}) and residual network architecture (\cite{he2016deep}) for cell segmentation. \cite{kraus2016} use multiple instance learning (MIL) to simultaneously segment and classify cells in microscopy images. The binary instance classifier generates the predictions which are combined through an aggregate function in the MIL layer of the proposed network. However, this approach can be computationally expensive for solving a segmentation problem as multiple feature maps need to be aggregated using the global pooling function. DCAN (\cite{Chen2017dcan}) employs a modified FCN that simultaneously segments both the objects and contours to assist separating clustered object instance. Another recently proposed multi-scale convolutional neural network (\cite{Song2016}) trains the network at different scales of the Laplacian pyramid and merges the network in the upsampling path to perform segmentation. \cite{xu2016gland, Xu2017} proposed a network that performs side supervision of boundary maps in addition to the foreground. \cite{Manivannan2017} combined handcrafted features with deep learning for segmentation, but this approach is computationally expensive as it not only requires calculation of features using classical approaches but also a support vector machine (SVM) classifier to predict local label patches.
	

	\section{Data Sets and Challenges}
	\label{sec:datasetsandchallenges}
	The data sets that we use in this paper come from two different sources. The first data set contains images acquired using a multiplexed fluorescence microscope, capable of acquiring images of multiple tags in a cyclic manner (\cite{schubert2006}), where our task was cell segmentation. The other two data sets are Haematoxylin and Eosin (H\&E) stained microscopic images collected as part of open challenge contests. We use one of the data sets to evaluate nuclear segmentation in four different tumour types (\cite{CPMwebsite}) and the other one for gland segmentation in colon cancer histology images (\cite{Sirinukunwattana2016GLaS}). In this way, we demonstrate that the proposed network
	is capable of dealing with diverse datasets and segmentation
	tasks.
	\subsection{Multiplexed Fluorescence Imaging Data}
	\label{sec:FluorImageData}
	We first focus on segmentation of individual cells in multiplexed fluorescence images using nuclear and membrane markers. In the fluorescence microscopy images, this task is challenging for various reasons, for example relatively large variation in intensity of captured signal and difficulty with separating neighbouring cells. It requires careful tuning of the algorithm to make it robust to intensity, shape, size and fusion of individual cellular regions. That process can require experimentation with a variety of features and can be time consuming. Membrane markers such as E-cadherin (or Ecad) mark the boundary of individual cells, but the intensity of the membrane markers varies depending on type and orientation of each cell which makes segmentation difficult. 
	
	A multi-channel fluorescence microscope known as the Toponome Imaging System (TIS) (\cite{schubert2006}), acquired images of tissue samples from mouse pancreata. The TIS microscope is capable of capturing signals from multiple biomarkers, but for cell segmentation we employ only two channels corresponding to Ecad (membrane marker using FITC channel) and DAPI (nuclear marker). After segmentation work is completed, the other channels are available to study individual cells, and to group similar cells together for statistical purposes. We performed alignment and normalization of the multi-channel images using protocols designed for pre-processing of the TIS data (\cite{Raza2012,Raza2016}). Next, ground truth for image segmentation, marked by an expert biologist, was used for training.
	
\begin{figure}[tbh!]
		\centering
		\includegraphics[width=0.8\textwidth]{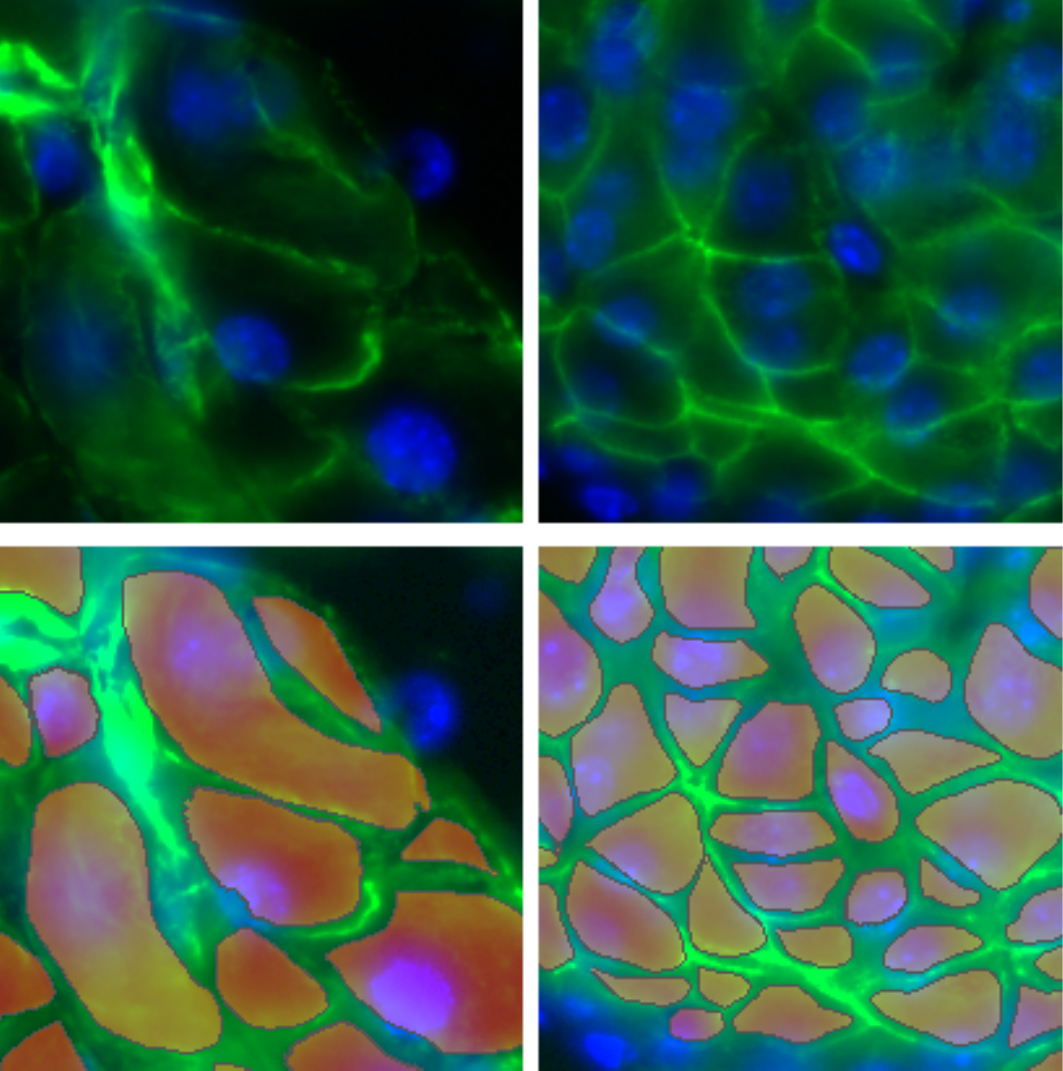}
		\caption{Top row: Membrane marker (Ecad-FITC) is shown in green and nuclear marker (DAPI) in blue. Bottom row: ground truth is overlaid in red with black boundaries. Left: Exocrine Cells. Right: Endocrine Cells.}
		\label{fig:SampleImages}
	\end{figure}

	Sample images of mouse pancreatic exocrine cells and endocrine cells are shown in Figure \ref{fig:SampleImages} as RGB composite images (enhanced for display), where membrane marker is shown in green, nuclear marker in blue and ground truth is overlaid in red with black boundaries. One can observe the variation in intensities of cell boundaries and that the nuclei are not always present and, if present, are not always positioned at the centre of the cell. This is because a tissue is a three-dimensional structure which is finely cut into multiple sections to obtain a two-dimensional image, which may or may not contain part of the cell containing the nucleus. Pancreatic cells are either endocrine cells, seen in the islets, or exocrine cells. Endocrine cells are more tightly packed and are smaller than exocrine cells. In addition, images with varying levels of signal-to-noise ratio (SNR) are expected in fluorescence microscopy images where not only the imaging apparatus but also antibody concentration, temperature and incubation times contribute to noise. These variations make segmentation a challenging task.
	
	\subsection{The Computational Precision Medicine (CPM) Data Set for nuclear segmentation}
	\label{sec:CPMData}
\begin{figure}[tbh!]
		\centering
		\includegraphics[width=0.8\textwidth]{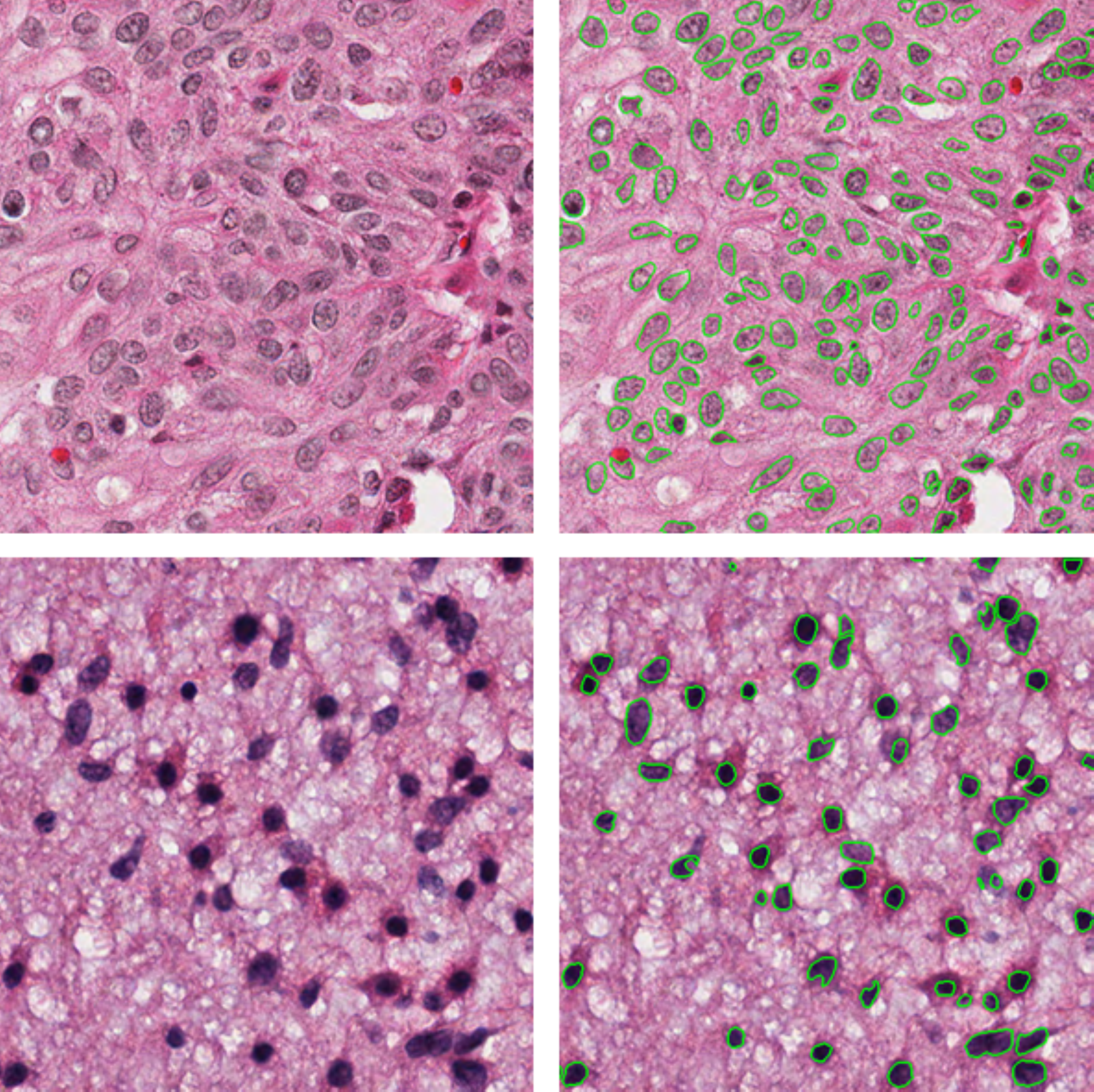}
		\caption{Left: Sample Images from the CPM data set. Right: Ground truth marking outlined in green colour on the sample images. Top row: head and neck squamous cell carcinoma. Bottom row: lower grade glioma.}
		\label{fig:SampleImagesCPM}
	\end{figure}

	Nuclear segmentation can help understand the tumour microenvironment by studying features such as nuclear pleomorphism and nuclear morphology. Segmentation of nuclei in histology images is difficult, especially within tumour cells due to their heterogeneous nature with high variation in shape, size and chromatin pattern. The data set we use in this paper was published as part of a challenge contest at Medical Image Computing and Computer Assisted Interventions (MICCAI) 2017. The data set contains 32 training and 32 testing image tiles along with ground truth marking for nuclear segmentation, extracted from multi-tissue H\&E stained histology slides. There is an equal representation of glioblastoma multiforme (GBM), lower grade glioma (LGG), head and neck squamous cell carcinoma (HNSCC) and non-small cell lung cancer (NSCLC). A couple of example images (left) with corresponding ground truth outlined with green boundary (right) are shown in Figure \ref{fig:SampleImagesCPM}.
	
	\subsection{Gland Segmentation (GLaS) Challenge Data Set}
	\label{sec:dataset}
\begin{figure}[tbh!]
		\centering
		\includegraphics[width=0.99\textwidth]{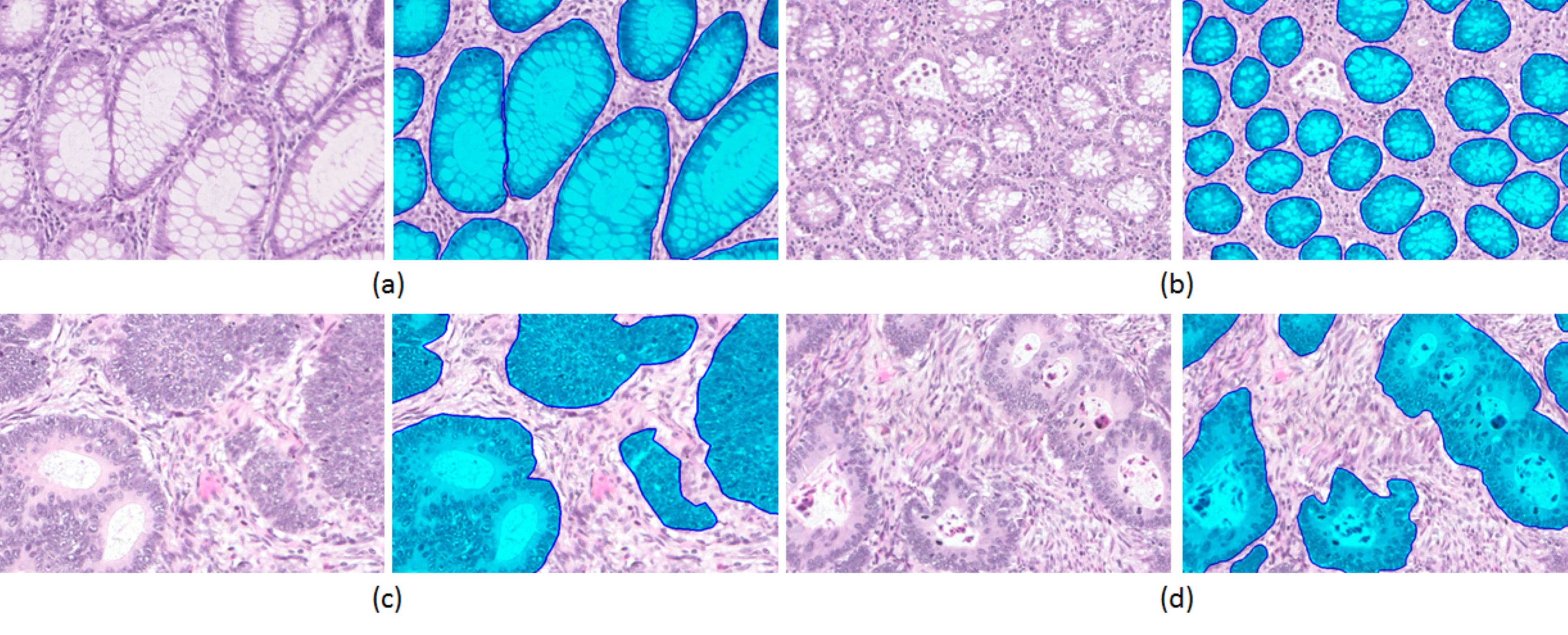}
		\caption{Sample images from the GLaS data set (\cite{Sirinukunwattana2016GLaS}). The images are shown in pairs, where the sample image on the left is overlaid on the right with the ground truth. The top row shows sample images from benign cases and the bottom row shows sample images from malignant cases. (a) \& (b) show variation in size and structure of glands in benign cases, whereas (c) \& (d) show variation in malignant colon cancer, where (c) is taken from a moderately differentiated sample and (d) is taken from a poorly differentiated (higher grade) cancerous sample.}
		\label{fig:sampleimageslabels}
	\end{figure}

	Histological assessment of glands is one of the key factors in colon cancer grading (\cite{Sirinukunwattana2016GLaS}). This requires a highly trained pathologist, is labour intensive, suffers from inter and intra-observer variability and has limited reproducibility. Due to complex nature of the problem, sophisticated algorithms are needed for successful automatic segmentation. Automatic segmentation of glands is challenging due to high variation in texture, size and structure of glands especially in malignant tissue. The third data set we use in this paper is the publicly available Warwick-QU data set published as part of the GLand Segmentation (GLaS) challenge (\cite{Sirinukunwattana2016GLaS}). The data set consists of 165 images with the associated ground truth marked by expert pathologists. The composition of the data set is detailed in Table \ref{table:datasetcomposition}, whereas a few sample images from the data set are shown in Figure \ref{fig:sampleimageslabels}. In Figure \ref{fig:sampleimageslabels}, the top row shows sample images from benign cases, and the bottom row shows sample images from malignant cases. Figure \ref{fig:sampleimageslabels} (c) has been taken from a moderately differentiated colon cancer tissue and (d) has been taken from a poorly differentiated colon cancer tissue section. It is evident from these images that there is a large variation in the size, texture and structure of glands in both malignant and benign cases although the variation is greater in malignant cases.
	
	\begin{table}[tbh!]
		\centering
		\caption{Composition of Warwick-QU data set.}
		\label{table:datasetcomposition}
		\begin{tabular}{|l|l|l|l|}
			\hline
			\multirow{2}{*}{Histologic Grade} & \multicolumn{3}{l|}{Number of images} \\ \cline{2-4}
			& Training     & Test A     & Test B    \\ \hline
			Benign                            & 37           & 33         & 4         \\ \hline
			Malignant                         & 48           & 27         & 16        \\ \hline
		\end{tabular}
	\end{table}

	\section{The Proposed Network}
	\label{sec:proposednetwork}
\begin{figure}[t!]
		\centering
		\includegraphics[width=1.0\textwidth]{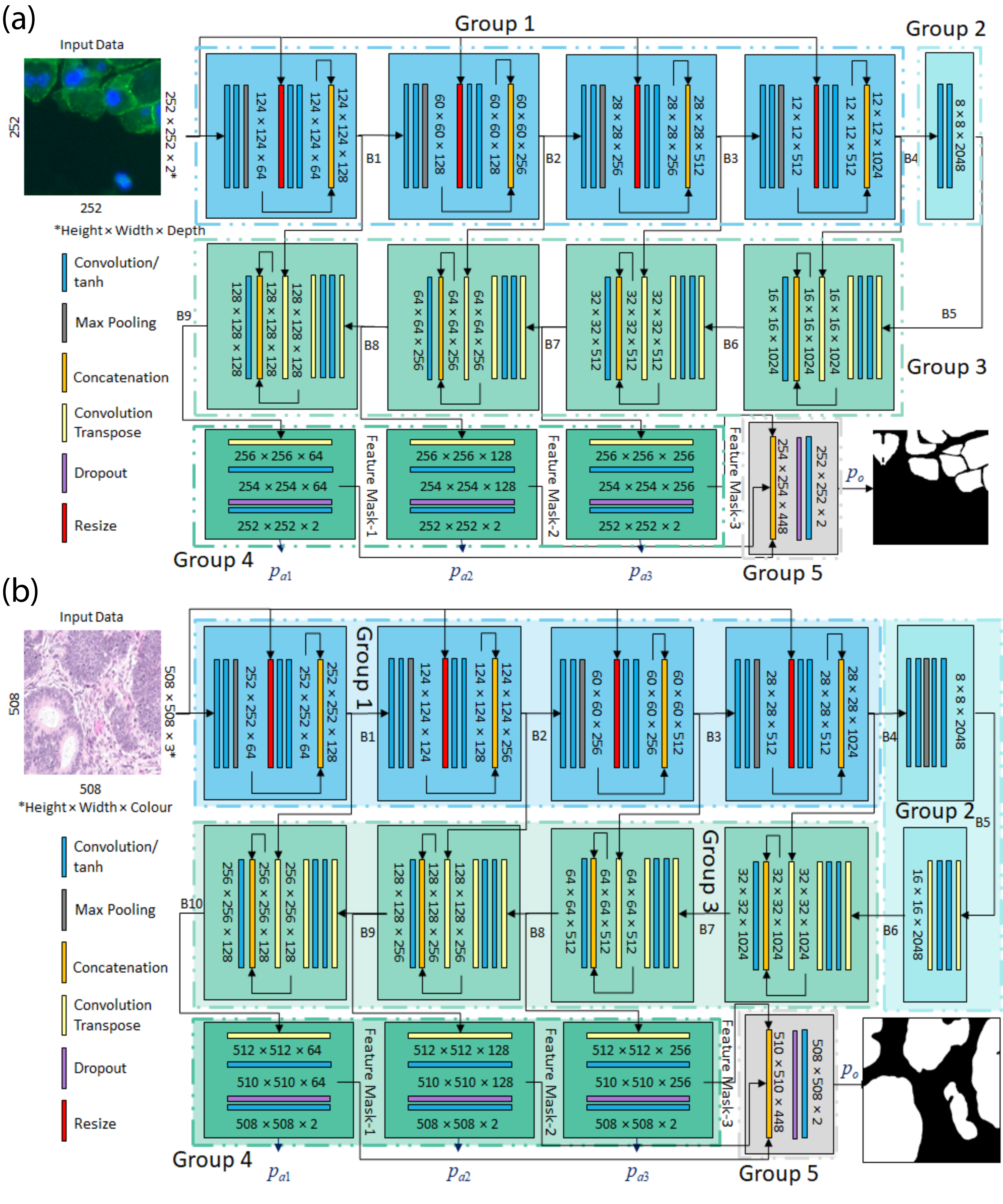}
		\caption{The proposed Micro-Net architecture. (a) Micro-Net-252 \& (b) Micro-Net-508.}
		\label{fig:NetworkArchitecture}
	\end{figure}

	The architecture of proposed Micro-Net is shown in Figure \ref{fig:NetworkArchitecture}. In the case of fluorescence images, the input to the network consists of two features, i.e., membrane and nuclear marker images, whereas in the case of H\&E images the input to the network is a stain normalised RGB image. We perform stain normalisation\footnote{\href{http://www2.warwick.ac.uk/fac/sci/dcs/research/tia/software/sntoolbox}{http://www2.warwick.ac.uk/fac/sci/dcs/research/tia/software/sntoolbox}} using the method proposed by (\cite{reinhard2001color}) to reduce the effect of stain variation from different labs and staining conditions. In both cases, the network performes batch normalisation at the input layer. The network is divided into five groups and thirteen branches, the division depending on their function and the set of layers/filters.
		
	
	
	\subsection{Group 1: Downsampling}
	The first group, which consists of four branches with output B1-B4, constructs the downsampling path. Each branch in Group 1 consists of convolution, max-pooling, resize and concatenation layers. The convolution and max-pooling layers perform standard operations as in conventional CNNs. We use \textit{tanh} activation after each convolution layer as our experiments showed that the network converges faster with \textit{tanh} activation than with ReLU. The resize layer resizes the image using bicubic interpolation so that the resized image dimension matches the corresponding dimension of the max-pooling output. We add the lower resolution input to retain the information from pixels that do not have the maximum response, because they are in the vicinity of a noisy neighbourhood. This is particularly useful when we are trying to retain tiny feature details ignored during the max-pooling operation, for example, when trying to detect cells with boundary markers having extreme intensities, even for individual cells as shown in Figure \ref{fig:SampleImages}. Another aspect of the resizing operation is to train the network on different sized cells/nuclei and glands as explained in Section \ref{sec:datasetsandchallenges}. The output of branch 1 (B1) has feature depth of size 128 where the first half (64) of the features are the result of the max-pooling operation and the next half (64) are obtained by performing convolutions only on the resized image. The following branches in Group 1 double the feature depth of the previous branch but follow the same protocol in generating the branch output. The only difference is that B1 performs batch normalisation at the input and the resize layer whereas B2-B4 perform batch normalisation at the resize layer only.
	
	\subsection{Group 2: Bridge}
	Group 2, consisting of B5, bridges the connection between the downsampling and upsampling paths, whose architecture is very similar to conventional CNN architectures. 
	
	\subsection{Group 3: Upsampling}
	Group 3 forms the upsampling path and consists of branches B6, B7, B8 \& B9. Each of these branches take two inputs, one from the previous branch and one from the branch with the closest feature dimension in the downsampling path. The output of each branch is double in height and width and half the depth of the previous branch. The second input is added from the downsampling path for better localization and to capture context information as in (\cite{Ronneberger2015}). It also passes the convolution only features to the upsampling path, which helps to learn from features which do not have maximum response in the downsampling path. Compared to U-Net (\cite{Ronneberger2015}), we add additional deconvolution layers instead of cropping the feature from the downsampling path. This allows us to produce a segmentation map of the same size as the input image and an overlap-tile strategy is not required. It also reduces the number of patches required to produce the desired segmentation output thus removing computational steps.
	
	\subsection{Group 4 \& 5: Auxiliary and Main Output}
	\label{Aux and Main}
	Group 4 \& 5 generate the auxiliary and main output and calculate the loss function. Group 4 consists of three branches where each branch takes output from one of B7-B9 and generates three auxiliary feature masks, which are fed into the main output branch. The output branch concatenates feature masks and performs convolution followed by softmax classification to get the segmentation output map $p_{o}(x)$ where $x$ represents a pixel location. The output of branches B7-B9 are of different resolutions and so the deconvolution  layer in each of the auxiliary branches is set to generate the output of the same size (\cite{Chen2017dcan}). The deconvolution is followed by a convolution layer which produces the auxiliary feature mask. Each of the auxiliary feature masks is followed by a dropout layer (set to 50\%) and the convolution layer followed by softmax classification to get the auxiliary outputs ($p_{a1}(x)$, $p_{a2}(x)$, $p_{a3}(x)$).
	
	\subsection{Modifications for Gland Segmentation}
	For gland segmentation we slightly modified the network to train on a bigger patch size as shown in Figure \ref{fig:NetworkArchitecture}(b). We doubled the input size to incorporate larger context to take account of the larger size of glands as compared individual cells. This modified architecture consists of five groups and fourteen branches where all five groups and the corresponding branches perform the same tasks as in Figure \ref{fig:NetworkArchitecture}(a). However, the architecture of group 2 was slightly modified to learn deep features by adding an additional branch that performs deconvolution followed by convolution. The additional branch in group 2 was added so that the smallest feature patch size is ($8\times8$) in line with the Micro-Net 252 architecture. The rest of the architecture remains the same except for the size of input/output for each branch.
	
	\subsection{Loss Function}
	\label{sec:lossfuntion}
	For training, we calculate weighted cross entropy loss for the main output ($l_o$) and the auxiliary outputs ($l_{a1}$, $l_{a2}$, $l_{a3}$) as
	\begin{align}
	{{l}_{k}}=\sum\limits_{x\in \Omega }{w(x)\log ({{p}_{k(x)}}(x))}
	\end{align}
	where $k\in\left\{o,a1,a2,a3\right\}$, as explained in Subsection~\ref{Aux and Main}  and $\Omega$ is the set of pixel locations in the input image. The weight function $w(x)$ gives higher weights to pixels that are at the merging cell boundaries, leading to a higher penalty (\cite{Ronneberger2015}). The total loss ($l$) is calculated by combining auxiliary and main loss by using $l = l_o + (l_{a1}+ l_{a2} + l_{a3})/epoch$
	where $epoch>0$ represents the number of training passes already made through the data. This strategy reduces exponentially the contribution of auxiliary losses for a higher $epoch$, avoiding reduction of the contribution by large steps (\cite{Chen2017dcan}).
	
	\subsection{Data Augmentation}
	As deep learning algorithms require large amounts of data for training, we augment the data using barrel, pincushion and moustache distortion. While adjusting parameters we made sure by visual examination that the distortions created by these parameters were realistic and not too strong. For cell segmentation on fluorescence imaging data, we augmented the data by adding white Gaussian noise with mean $0$ and variance in the range $0.0007$ to $0.001$, where for each patch the value of variance was randomly selected. For nuclear and gland segmentation we introduced Gaussian blur with a Gaussian filter of size $12 \times 12$, with $\sigma$ ranging from 0.2 to 2. The value $\sigma$ was randomly selected for each patch. In addition we rotate, and flip the images left, right, up and down. To train the network for cell/nuclear (gland) segmentation, we first extract $300 \times 300$ ($600 \times 600$) patches from the training data. If the size of image is smaller than $300$ ($600$) in height or width, we symmetrically pad the image to increase its size. During training the network picks these patches in a random order for each $\mathit{epoch}$, choosing centres for the patches at random locations, and then cropping them to a size of $252 \times 252$ ($508 \times 508$) patch before inputting. The proposed network was implemented using TensorFlow v0.12 (\cite{tensorflow2015-whitepaper}). We start with a learning rate ($lr = 0.001$) and reduce it according to $lr=0.001/(10^{(\mathit{epoch}/5)})$, which reduces the learning rate by a factor of 10 for every fifth $\mathit{epoch}$.
	
	\section{Results and Discussion}
	\label{sec:results}
		
	\subsection{Multiplexed Fluorescence Imaging Data}
	\label{sec:FluorImageDataResults}
	\begin{figure}[tbh]
		\centering
		\includegraphics[width=0.99\textwidth]{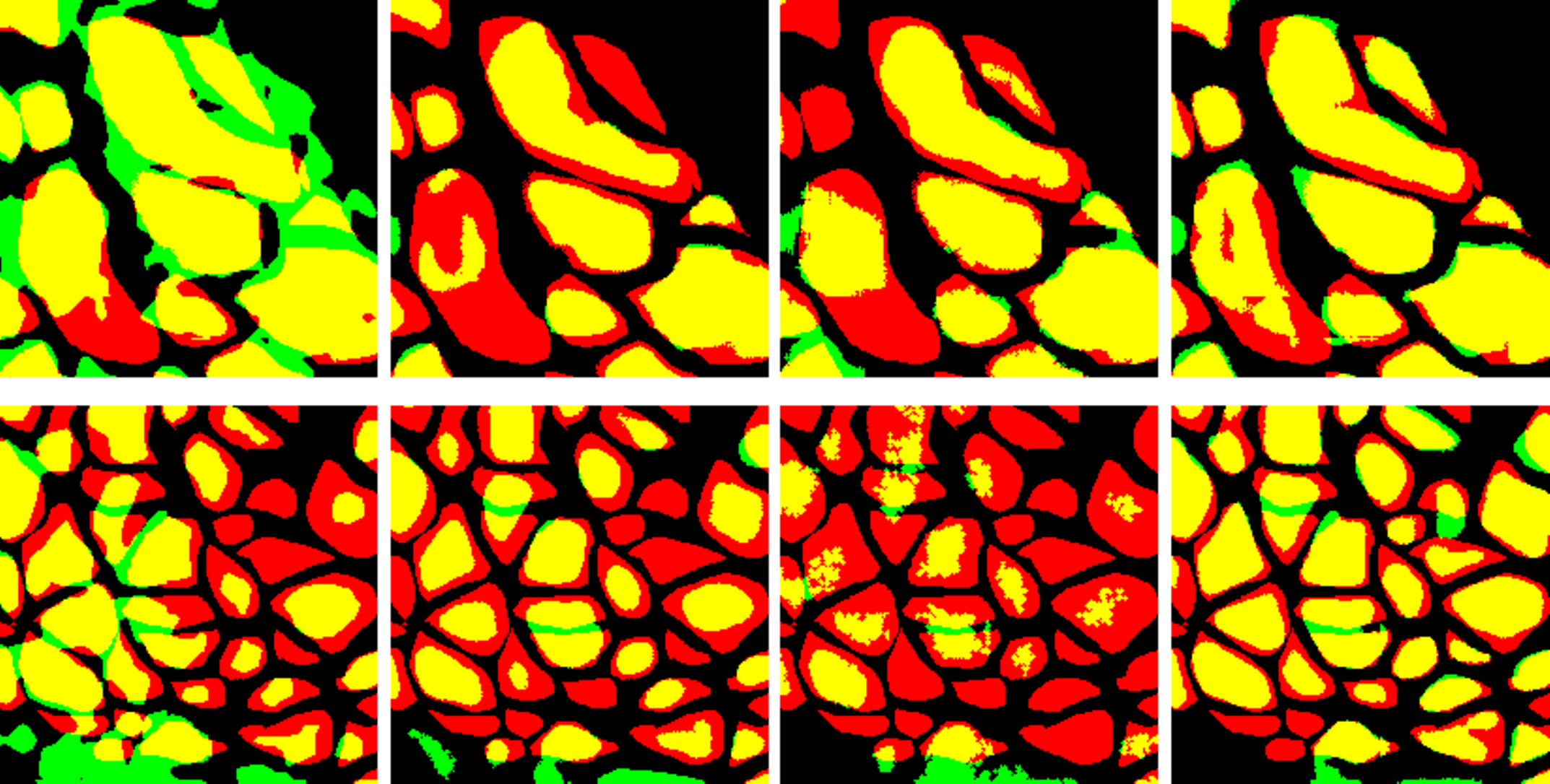}
		\caption{Segmentation results, ground truth in red, output of the algorithm in green and overlap between ground truth and output of algorithm in yellow. Top row: Exocrine region. Bottom row: Endocrine region. Columns (left to right) are output from FCN8, U-Net, DCAN and Micro-Net architectures.}
		\label{fig:Results}
	\end{figure}
	Our image data consists of 10 images of size $2048 \times 2048$ pixels (11,163 cells) of which 6 images (with approximately 60\% i.e., 6,641 cells) are used for training and 4 images (with the remaining 40\% i.e., 4,522 cells) for testing. During training, we used 20\% of the data for validation. We compare our results with the state-of-the-art FCN8 (\cite{FCN2016}), DCAN (\cite{Chen2017dcan}) and U-Net (\cite{Ronneberger2015}) networks. To remove the bias we trained all the networks on the same training data obtained after augmentation. We used the authors' implementation of FCN8 and trained it for our data, whereas DCAN and U-Net were implemented in TensorFlow. The weights for the proposed network were initialised with truncated Gaussian and the network was trained for $25$ epochs. The checkpoint was chosen based on the minimum validation loss. For U-Net and DCAN, we ran the network for $30$ epochs but the criteria for choosing the trained checkpoint file was the same (i.e., best checkpoint was based on minimum validation loss.) The results (Figure \ref{fig:Results}) show that FCN8 identified cellular regions but was not able to segment individual cells. DCAN is designed to learn the contour features and performed better segmentation of the cells in the exocrine region but performed poorly with smaller sized cells in the endocrine region. U-Net performed better than both FCN and DCAN but missed the cells with weaker boundaries. The proposed Micro-Net method, performed better in the presence of variable intensities and variable size/shape of the cells. The output in Figure \ref{fig:Results} was post-processed for all the algorithms using area opening (100 pixels) and hole filling operations to get the final output score in Table \ref{table:1}. For quantitative analysis, we used measures which include Dice coefficient, F1 score, object Dice, pixel accuracy and object Hausdorff (\cite{Sirinukunwattana2016GLaS}). Better results correspond to smaller Hausdorff distance and all other measures larger. The quantitative results are shown in Table \ref{table:1} which show that the proposed Micro-Net method outperforms the state-of-the-art deep learning approaches with at least 3-4\% margin in terms of average Dice, F1 score, object Dice, pixel accuracy and object Hausdorff. We modified the FCN8 algorithm (FCN8W) by introducing weighted loss (\cite{Ronneberger2015}) to improve segmentation of individual cells. FCN8W improved F1, object Dice, pixel accuracy and object Hausdorff but failed to increase the Dice coefficient. In addition, we add results for Micro-Net\textsuperscript{--} where we removed the multi-resolution input and the bypass layers. The results are slightly better compared to U-Net but not better than Micro-Net supporting the suggested changes in the design.
		
	\begin{table}[t!]
		\centering
		\caption{Quantitative results for cell segmentation in terms of Dice coefficient, F1 score, Object Dice (OD), Pixel Accuracy (Acc) \& Object Hausdorff (OH).}
		\label{table:1}
		\begin{tabular}{l|c|c|c|c|c}
			\hline
			\hline
			Network                                            & Dice             & F1               & OD               & PAcc             & OH             \\ \hline
			\begin{tabular}[c]{@{}l@{}}FCN8\\ (\cite{FCN2016})\end{tabular}  & 78.54\%          & 11.69\%          & 7.49\%           & 77.68\%          & 1349.51        \\ \hline
			FCN8W                                              & 71.36\%          & 50.53\%          & 50.86\%          & 74.61\%          & 91.77          \\ \hline
			\begin{tabular}[c]{@{}l@{}}DCAN\\ (\cite{Chen2017dcan})\end{tabular}  & 76.03\%          & 61.41\%          & 63.80\%          & 78.67\%          & 42.31          \\ \hline
			\begin{tabular}[c]{@{}l@{}}U-Net\\ (\cite{Ronneberger2015})\end{tabular} & 78.39\%          & 66.43\%          & 67.35\%          & 80.28\%          & 40.49          \\ \hline
			{Micro-Net\textsuperscript{--}}           & 80.74\%         & 69.87\%         & 71.52\%          & 82.22\%        & 30.61       \\ \hline
			Proposed                                           & \textbf{82.43\%} & \textbf{71.79\%} & \textbf{74.12\%} & \textbf{83.53\%} & \textbf{27.53} \\ \hline
		\end{tabular}
	\end{table}

	In addition to the above experiments we tested all the network architectures for their robustness to various levels of white Gaussian noise by controlling the SNR and generating the output for various network architectures. For this purpose we did not retrain the networks but used the already trained models as above. The values for Dice, F1, object Dice, pixel accuracy and object Hausdorff for various SNR values are given in Table \ref{table:SNRDICE}, \ref{table:SNRF1}, \ref{table:SNRODICE}, \ref{table:SNRPAcc} and \ref{table:SNROH} respectively. For Dice coefficient, FCN8 and U-Net drop to 73\% whereas the proposed method drops to 78\%. However, DCAN shows a different behaviour and increases the Dice values by 2\% taking the top spot at 1dB. In terms of the F1 score, all the networks drop to below 65\% at 1dB where the proposed network retains the top position. The trend was similar with object Dice as well where the first position was retained by the proposed network. In terms of pixel accuracy, the proposed method dropped roughly 4\% from 20dB to 1dB. DCAN however showed similar trend to Dice values i.e., increased pixel accuracy with increasing noise content. This is an interesting result, but if we carefully observe Dice and pixel accuracy both depend only on the pixels that are in agreement, whereas object Dice only takes into account the pixels which belong to the ground truth. Similarly, the F1 score takes into account not only true positives but also false positives. With increasing noise content, DCAN lost some of the false positives causing more pixels not belonging to the cellular region to agree. This increased pixel accuracy and Dice while simultaneously decreasing object Dice and the F1 score. Object Hausdorff is a measure of similarity of the shape of cell boundaries and is lower for better performance. Here the author's implementation of FCN produced very high values, whereas with weighted loss (FCN8W) the values were in a comparable range. Additionally FCN8W values improved with increasing noise due to the resulting loss of many cell segmentations. U-Net increased 55 points from 20dB to 1dB noise which shows it's sensitivity to noise. DCAN performed better as it is designed to match the contours and only increased 12 points with increased noise content.  However, the proposed network only increased 6 points showing the stability of the network and it's robustness to the noise. It is important to note here that Micro-Net\textsuperscript{--} shows the steepest decline in performance (in terms of Dice, F1, object Dice and pixel accuracy) with increasing noise content compared to U-Net, DCAN and Micro-Net. The decline is not much in the case of object Hausdorff as Micro-Net\textsuperscript{--} uses the multi-resolution output, and therefore shows robust shape similarity (on segmented cells) as shown by DCAN (which also utilises multi-resolution output). This further emphasises the role of multi-resolution input and the bypass layers. 
	
	Overall, the proposed network outperformed the recent deep learning models for various levels of noise in terms of Dice, the F1 score, object Dice, pixel accuracy and object Hausdorff. In terms of computation time, on average the network takes 2.50 sec/batch for training and 0.39 sec/batch for testing a batch of $5$ images on a Windows10 machine with Intel Xeon E5-2670 v2 CPU, TitanX Maxwell GPU and 96 GB RAM. For $25$ epochs, it took around $4.5$ days to train the network.
	
	\begin{table}[tbh!]
		\centering
		\caption{Quantitative results for cell segmentation of state-of-the-art algorithms in terms of Dice coefficient for various SNR values.}
		\label{table:SNRDICE}
		\begin{tabular}{l|l|l|l|l|l|l}
			\hline
			\hline
			Network                                            & 20dB             & 15dB             & 10dB             & 5dB              & 3dB              & 1dB              \\ \hline
			\begin{tabular}[c]{@{}l@{}}FCN8\\ (\cite{FCN2016})\end{tabular}  & 77.55\%          & 76.24\%          & 76.97\%          & 76.47\%          & 75.26\%          & 73.66\%          \\ \hline
			FCN8W                                              & 71.56\%          & 71.53\%          & 71.39\%          & 70.56\%          & 69.36\%          & 66.88\%          \\ \hline
			\begin{tabular}[c]{@{}l@{}}DCAN\\ (\cite{Chen2017dcan})\end{tabular}  & 77.18\%          & 77.95\%          & 78.92\%          & 79.82\%          & 79.91\%          & \textbf{79.40\%} \\ \hline
			\begin{tabular}[c]{@{}l@{}}U-Net\\ (\cite{Ronneberger2015})\end{tabular} & 76.58\%          & 76.61\%          & 76.29\%          & 75.59\%          & 74.99\%          & 73.45\%          \\ \hline
			{Micro-Net\textsuperscript{--}}          & 81.35\%          & 81.42\%          & 80.72\%          & 78.30\%          & 76.04\%          & 72.33\%          \\ \hline
			Proposed                                           & \textbf{82.62\%} & \textbf{82.69\%} & \textbf{82.49\%} & \textbf{81.37\%} & \textbf{80.23\%} & 78.52\%          \\ \hline
		\end{tabular}
	\end{table}
	
	\begin{table}[tbh!]
		\centering
		\caption{Quantitative results for cell segmentation of state-of-the-art algorithms in terms of F1 score for various SNR values.}
		\label{table:SNRF1}
		\begin{tabular}{l|l|l|l|l|l|l}
			\hline
			\hline
			Network                                            & 20dB             & 15dB             & 10dB             & 5dB              & 3dB              & 1dB              \\ \hline
			\begin{tabular}[c]{@{}l@{}}FCN8\\ (\cite{FCN2016})\end{tabular}  & 16.69\%          & 26.34\%          & 15.25\%          & 5.34\%           & 4.33\%           & 4.06\%           \\ \hline
			FCN8W                                              & 49.41\%          & 49.54\%          & 49.07\%          & 47.75\%          & 48.24\%          & 44.92\%          \\ \hline
			\begin{tabular}[c]{@{}l@{}}DCAN\\ (\cite{Chen2017dcan})\end{tabular}  & 62.33\%          & 63.15\%          & 63.65\%          & 62.18\%          & 61.58\%          & 60.16\%          \\ \hline
			\begin{tabular}[c]{@{}l@{}}U-Net\\ (\cite{Ronneberger2015})\end{tabular} & 65.51\%          & 65.01\%          & 65.05\%          & 63.18\%          & 61.19\%          & 58.27\%          \\ \hline
			{Micro-Net\textsuperscript{--}}          & 70.43\%  & 69.99\%          & 68.95\%          & 66.07\%          & 63.23\%          & 58.68\%          \\ \hline
			Proposed                                           & \textbf{71.63\%} & \textbf{71.59\%} & \textbf{70.65\%} & \textbf{69.19\%} & \textbf{67.93\%} & \textbf{64.67\%} \\ \hline
		\end{tabular}
	\end{table}

	\begin{table}[tbh!]
		\centering
		\caption{Quantitative results for cell segmentation of state-of-the-art algorithms in terms of Object Dice for various SNR values.}
		\label{table:SNRODICE}
		\begin{tabular}{l|l|l|l|l|l|l}
			\hline
			\hline
			Network                                            & 20dB             & 15dB             & 10dB             & 5dB              & 3dB              & 1dB              \\ \hline
			\begin{tabular}[c]{@{}l@{}}FCN8\\ (\cite{FCN2016})\end{tabular}  & 12.30\%          & 23.47\%          & 13.26\%          & 5.27\%           & 4.55\%           & 4.50\%           \\ \hline
			FCN8W                                              & 50.78\%          & 50.77\%          & 50.57\%          & 50.28\%          & 50.97\%          & 51.01\%          \\ \hline
			\begin{tabular}[c]{@{}l@{}}DCAN\\ (\cite{Chen2017dcan})\end{tabular}  & 64.46\%          & 65.36\%          & 65.54\%          & 64.22\%          & 62.99\%          & 61.61\%          \\ \hline
			\begin{tabular}[c]{@{}l@{}}U-Net\\ (\cite{Ronneberger2015})\end{tabular} & 65.61\%          & 65.04\%          & 64.42\%          & 61.89\%          & 60.37\%          & 58.55\%          \\ \hline
			{Micro-Net\textsuperscript{--}}       & 71.60\%          & 71.28\%         & 70.57\%         & 68.28\%          & 66.45\%          & 63.11\%          \\ \hline
			Proposed                                           & \textbf{73.90\%} & \textbf{73.79\%} & \textbf{72.98\%} & \textbf{71.70\%} & \textbf{70.42\%} & \textbf{68.03\%} \\ \hline
		\end{tabular}
	\end{table}
	
	\begin{table}[tbh!]
		\centering
		\caption{Quantitative results for cell segmentation of state-of-the-art algorithms in terms of pixel accuracy for various SNR values.}
		\label{table:SNRPAcc}
		\begin{tabular}{l|l|l|l|l|l|l}
			\hline
			\hline
			Network                                            & 20dB             & 15dB             & 10dB             & 5dB              & 3dB              & 1dB              \\ \hline
			\begin{tabular}[c]{@{}l@{}}FCN8\\ (\cite{FCN2016})\end{tabular}  & 77.33\%          & 77.54\%          & 77.50\%          & 75.19\%          & 73.63\%          & 72.14\%          \\ \hline
			FCN8W                                              & 74.94\%          & 74.92\%          & 74.80\%          & 74.10\%          & 73.24\%          & 71.67\%          \\ \hline
			\begin{tabular}[c]{@{}l@{}}DCAN\\ (\cite{Chen2017dcan})\end{tabular}  & 79.57\%          & 80.18\%          & 80.93\%          & 81.42\%          & 81.30\%          & \textbf{80.55\%} \\ \hline
			\begin{tabular}[c]{@{}l@{}}U-Net\\ (\cite{Ronneberger2015})\end{tabular} & 78.64\%          & 78.65\%          & 78.30\%          & 77.40\%          & 76.80\%          & 75.28\%          \\ \hline
			{Micro-Net\textsuperscript{--}}          & 82.63\%          & 82.65\%          & 82.11\%          & 80.37\%          & 78.83\%          & 76.39\%          \\ \hline
			Proposed                                           & \textbf{83.69\%} & \textbf{83.71\%} & \textbf{83.48\%} & \textbf{82.49\%} & \textbf{81.52\%} & 80.10\%          \\ \hline
		\end{tabular}
	\end{table}

	\begin{table}[tbh!]
		\centering
		\caption{Quantitative results for cell segmentation of state-of-the-art algorithms in terms of object Hausdorff for various SNR values.}
		\label{table:SNROH}
		\begin{tabular}{l|l|l|l|l|l|l}
			\hline
			\hline
			Network                                            & 20dB           & 15dB           & 10dB           & 5dB            & 3dB            & 1dB            \\ \hline
			\begin{tabular}[c]{@{}l@{}}FCN8\\ (\cite{FCN2016})\end{tabular}  & 1171.08        & 604.18         & 1176.88        & 1618.90        & 1700.83        & 1589.12        \\ \hline
			FCN8W                                              & 88.55          & 90.42          & 86.99          & 84.24          & 81.03          & 66.30          \\ \hline
			\begin{tabular}[c]{@{}l@{}}DCAN\\ (\cite{Chen2017dcan})\end{tabular}  & 42.20          & 41.16          & 42.32          & 47.72          & 50.92          & 54.21          \\ \hline
			\begin{tabular}[c]{@{}l@{}}U-Net\\ (\cite{Ronneberger2015})\end{tabular} & 55.53          & 63.53          & 62.50          & 85.48          & 97.49          & 100.15         \\ \hline
			{Micro-Net\textsuperscript{--}}          & 31.42          & 32.13          & 32.85          & 33.62          & 33.91          & 35.13          \\ \hline
			Proposed                                           & \textbf{27.98} & \textbf{28.35} & \textbf{29.64} & \textbf{30.73} & \textbf{31.69} & \textbf{33.84} \\ \hline
		\end{tabular}
	\end{table}

	\subsection{Computational Precision Medicine (CPM) Data Set}
	\label{sec:CPMDataResults}
\begin{figure}[tbh!]
		\centering
		\includegraphics[width=0.95\textwidth]{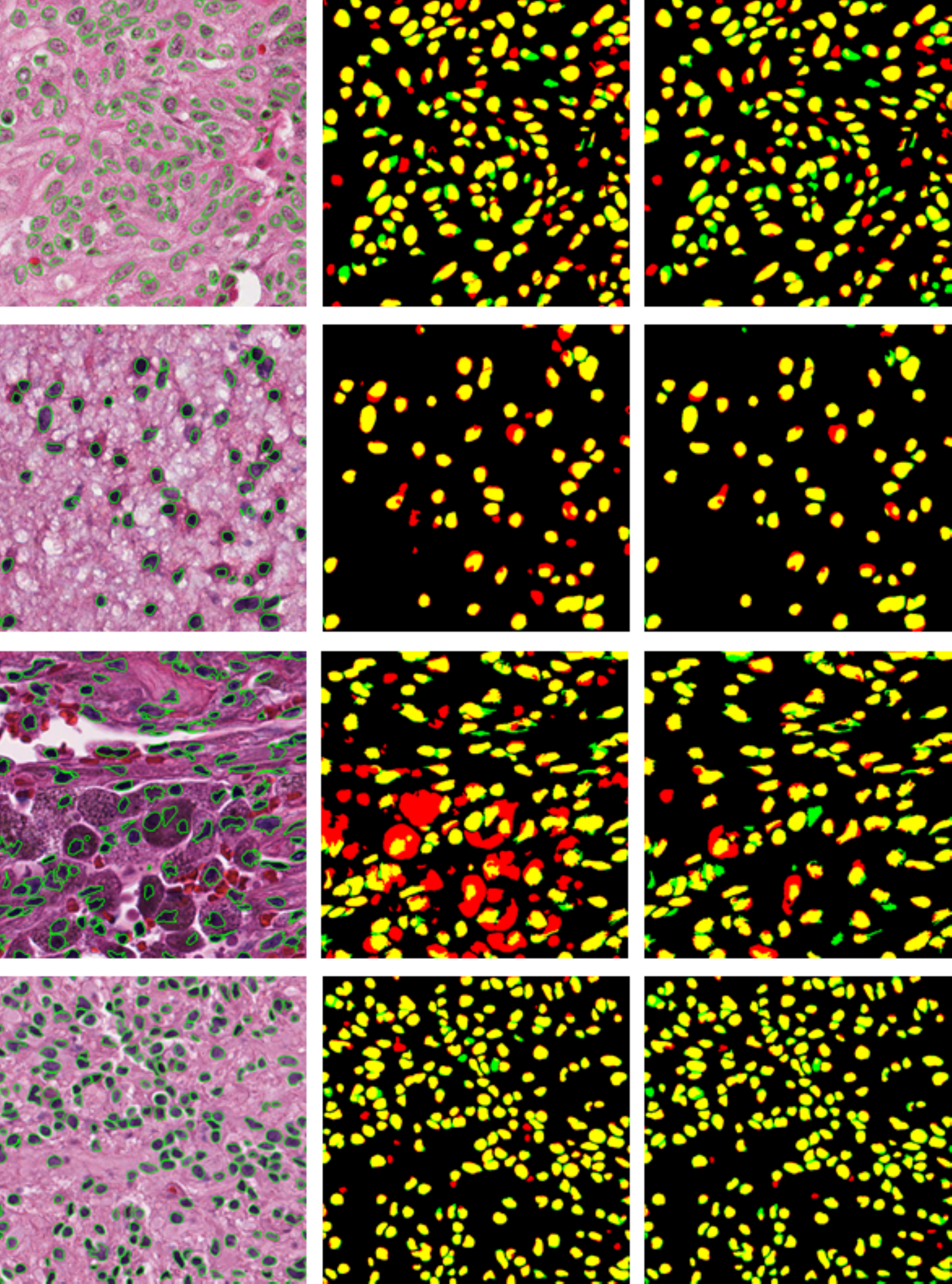}
		\caption{Top to Down: (1) HNSCC, (2) LGG, (3) NSCLC, (4) GBM. Left to Right: (1) H \& E image with ground truth outlined for nuclear segmentation, (2) RGB composite image with ground truth in green and output of U-Net in red, (3) RGB composite image with ground truth in green and output of the proposed algorithm in red. Yellow is the overlap between the ground truth and the output of the algorithm.}
		\label{fig:Results_CPM}
	\end{figure}

	Figure \ref{fig:Results_CPM} shows the performance of the proposed method on CPM data set. Row 1-4 show sample images from head and neck squamous cell carcinoma (HNSCC), lower grade glioma (LGG), non-small cell lung cancer (NSCLC) \& glioblastoma multiforme (GBM) respectively. Column 1 shows H \& E image with ground truth outlined for nuclear segmentation. Column 2 shows an RGB composite image with ground truth in green and output of U-Net in red and overlap in yellow. Column 3 is similar to column 2. It shows an RGB composite image with ground truth in green, the output of the proposed algorithm in red and the overlap in yellow. In HNSCC, on the lower left corner we can observe red cells, that are segmented by the proposed algorithm, but not picked during ground truth marking. On the other hand, on the top right, the algorithm undersegments, due to the relatively weak signal of the nuclei. In LGG, the proposed algorithm seems to oversegment where the ground truth might have been mistakenly marked on the nucleolus rather than on the nucleus of the cell. The algorithm seems to be struggling in these kinds of regions in NSCLC as well, where it missed segmenting the nuclei under darker shades. In GBM, the proposed algorithm misses a few cells with fainter nuclei. Overall the performance of the algorithm seems good; however it seems to struggle with nuclei under multiple shades. U-Net on the other hand shows more red in all cases demonstrating oversegmentation. This is particularly noticeable in NSCLC where there seems to be a strong cytoplasmic shade. Micro-Net struggles in these circumstances but performs much better than U-Net due to its robustness. 
	We quantitatively measure the performance of segmentation using two evaluation metrics as selected by the contest organisers (\cite{CPMwebsite}), namely Traditional Dice (Dice 1) and Ensemble Dice (Dice 2). Dice 1 measures the
	overlap between the ground truth and the prediction, whereas
	Dice 2 also penalises the prediction if there is a mismatch in the way segmentation regions are split. The overall score is then computed as the average of the two Dice coefficients. We compare our results with FCN8 (\cite{FCN2016}), U-Net (\cite{Ronneberger2015}), SAMS-Net (\cite{Graham2018}) and the submissions in the competition. The results in Table \ref{tab:CPMResults} show that our method not only outperforms the results of contest winners but also recent deep learning methods such as SAMS-Net (\cite{Graham2018}).
	
	\begin{table}[tbh!]
		\centering
		\caption{Cell segmentation results on CPM contest data set using evaluation metrics Dice 1 and Dice 2 as selected by the organisers.}
		\label{tab:CPMResults}
		\begin{tabular}{l|l|l|l|l}
			\hline
			\hline
			Method     & Dice 1         & Dice 2         & Score          & Rank       \\ \hline
			Proposed   & \textbf{0.857} & \textbf{0.796} & \textbf{0.827} & \textbf{1} \\ \hline
			SAMS-Net (\cite{Graham2018})   & 0.855          & 0.769          & 0.812          & 2          \\ \hline
			U-Net (\cite{Ronneberger2015})     & 0.837          & 0.741          & 0.789          & 3          \\ \hline
			vuquocdang & -              & -              & 0.783          & 4          \\ \hline
			brisker    & -              & -              & 0.773          & 5          \\ \hline
			FCN8 (\cite{FCN2016})       & 0.829          & 0.697          & 0.763          & 6          \\ \hline
			Schwarz    & -              & -              & 0.703          & 7          \\ \hline
		\end{tabular}
	\end{table}
	
	\subsection{Gland Segmentation Challenge (GLaS) Data Set}
	\label{sec:GLaSdatasetResults}
\begin{figure}[t!]
		\centering
		\includegraphics[width=0.95\textwidth]{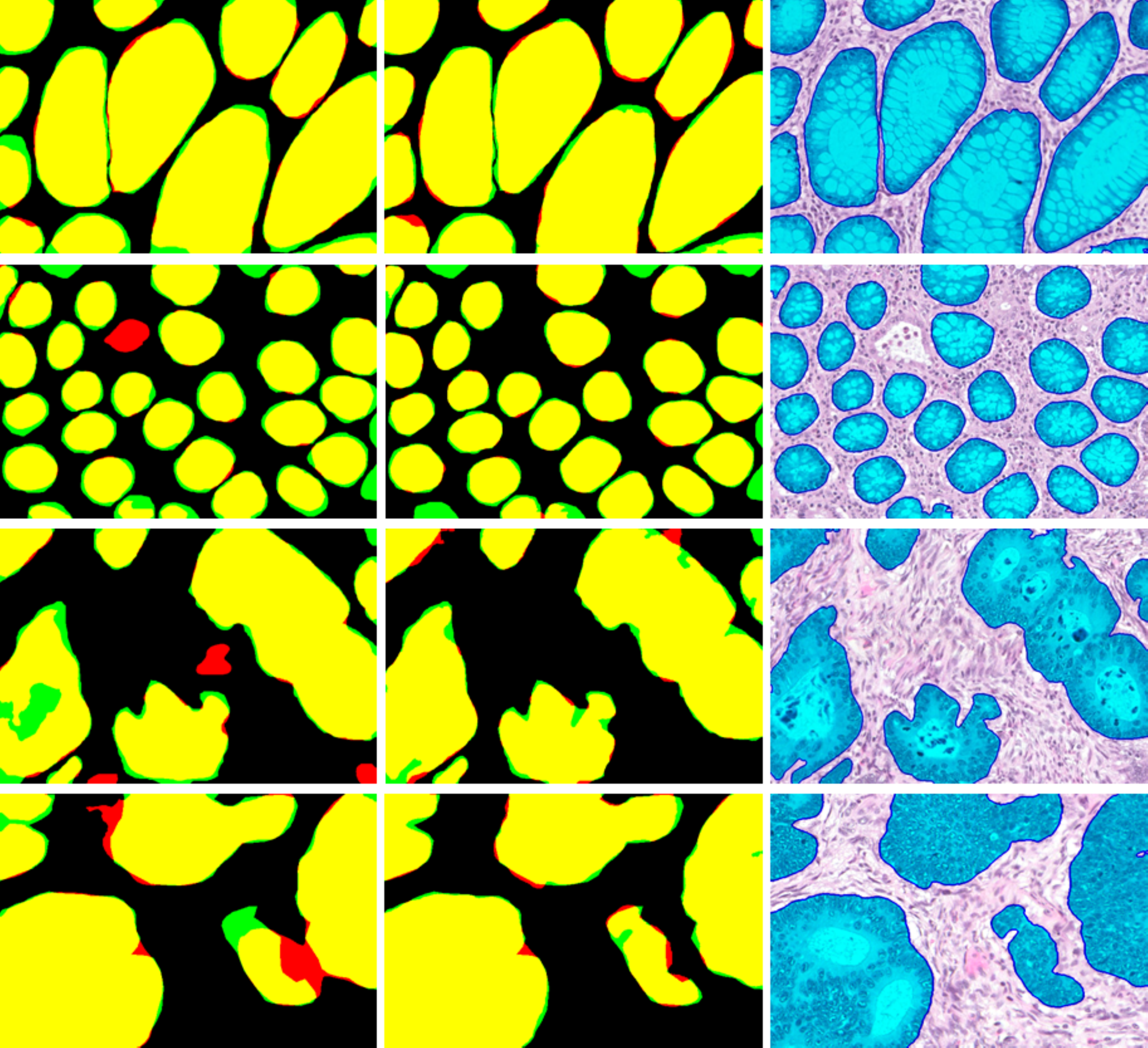}
		\caption{Results of the proposed network for sample images in Figure \ref{fig:sampleimageslabels}. First and second column show RGB composite images with output of the DCAN (\cite{Chen2017dcan}) and the proposed algorithm respectively in red, ground truth in green and overlap of ground truth and output in yellow. The image in the third column shows the output of the proposed method overlaid on the sample image. The results for DCAN were obtained from the contest organisers.}
		\label{fig:results}
	\end{figure}

	We used 85 images for training and 80 for testing, where 60 of the test images correspond to Test set A and the remaining 20 to Test set B. For quantitative analysis, we used Dice coefficient, F1 score, object Dice, pixel accuracy and object Hausdorff (\cite{Sirinukunwattana2016GLaS}). In the case of Hausdorff distance lower values are better; for other measures  higher are better. The quantitative results are given in Table \ref{table:results} which shows that our method produces competitive results compared to the state-of-the-art algorithms from the contest and ranks third after the recently proposed (\cite{Xu2017, Manivannan2017}) according the rank sum criteria set by the organisers. \cite{Xu2017} and \cite{Manivannan2017} employ optimised handcrafted features in addition to deep learning. \cite{Xu2017} on the other hand is heavily optimised for gland segmentation. Our algorithm employs only deep learning and is being proposed as a generic approach for cell, nuclear and gland segmentation, easily implemented using existing libraries optimised for GPU usage, which reduces the computational cost. Nevertheless, the proposed method performs competitively against recently proposed approaches and beats the conference version of \cite{xu2016gland}.  Compared to the results of the contest on test A, the proposed algorithm performed best in terms of F1 and object Dice but ranked third and fifth on test B. In terms of object Hausdorff, which measures the shape similarity, it ranked second after CUMedVision2 (\cite{Chen2017dcan}) on test A and \cite{xu2016gland} on test B. Lower F1 and object Dice on test B suggests that our method missed more glands in malignant cases, whereas lower Hausdorff suggests higher shape similarity to the ground truth extracted by our method. Qualitative results of the algorithm for sample images in Figure \ref{fig:sampleimageslabels} are shown in Figure \ref{fig:results}, where for each pair the image on first and second column show ground truth in green, output of DCAN (contest winner, the results were obtained from the contest organisers) and the proposed algorithm respectively in red and overlap of ground truth and output of algorithm in yellow. The image on the right shows the output of the proposed algorithm overlaid on the sample image. These results show that our algorithm clearly misses a few  glands on the boundary of the image for which there is insufficient information. In the second row, it merges the glands at the bottom of the image and misses one gland. In malignant cases the algorithm seems to be rather `conservative' in its approach when marking the boundary of glands. It can be observed in the third row for the two large glands at the bottom and in the fourth row for the smaller gland in the middle. All these glands show significant green inside the ground truth boundary, which at first suggests that the algorithm segmented the gland well inside the ground truth marking for the gland. However, when carefully observed in the overlay with the sample images the algorithm is faithfully following the boundary with tumor cells. For the large gland on the top right in third row the algorithm `oversegments' the gland compared to the ground truth but again, looking at the overlay with the sample image, the algorithm has included tumor cells in the segmentation. Overall the algorithm performs a good job in segmenting the glands but needs to improve on the glands at the boundary of a patch. This limitation could be overcome by using overlapping patches from the whole slide and then merging the results. Compared to DCAN (first column), the proposed algorithm shows better overlap with the ground truth. It can be observed that DCAN is sensitive to white spaces and certain architecture in benign cases where it segments false regions. This can be clearly observed in cases from row 2, 3 \& 4 and column 1. In the fourth row, it joins two glands together and under segments the smaller gland in the middle. It can also be observed that similar to the proposed algorithm, DCAN misses the glands at the boundary due to insufficient information. Overall the proposed algorithm performs better in terms of qualitative and quantitative results.
	
	\begin{sidewaystable}[ph!]
		\centering
		\caption{Quantitative comparison with the state-of-the-art methods. S and R in the table correspond to score and rank. It is important to note that \cite{Xu2017} and \cite{Manivannan2017} employ optimised handcrafted features in addition to deep learning.  \cite{Xu2017} on the other hand is optimised for gland segmentation. The proposed generic CNN framework managed to compete with recently proposed algorithms optimised for gland segmentation which demonstrates its robustness.}
		\label{table:results}
		\begin{tabular}[width=0.99\textwidth]{l|c|c|c|c|c|c|c|c|c|c|c|c|c}
			\hline
			\hline
			\multirow{3}{*}{Method} & \multicolumn{4}{c|}{F1 score}                             & \multicolumn{4}{c|}{Object Dice}                             & \multicolumn{4}{c|}{Object Hausdorff}                        & \multirow{3}{*}{Rank Sum} \\ \cline{2-13}
			& \multicolumn{2}{c|}{Test A} & \multicolumn{2}{c|}{Test B} & \multicolumn{2}{c|}{Test A} & \multicolumn{2}{c|}{Test B} & \multicolumn{2}{c|}{Test A} & \multicolumn{2}{c|}{Test B} &                           \\ \cline{2-13}
			& S             & R            & S            & R           & S             & R           & S            & R            & S              & R          & S          & R             &                           \\ \hline
			\cite{Xu2017}       & 0.893         & 4            & \textbf{0.843}        & \textbf{1}           & \textbf{0.908}         & \textbf{1}           & 0.833         & 2           & \textbf{44.129}        & \textbf{1}          & 116.821         & 2          & \textbf{11}                        \\ \hline
			\cite{Manivannan2017}   & 0.892         & 5            & 0.801        & 2           & 0.887         & 5           & \textbf{0.853}         & \textbf{1}        & 51.175         & 4          & \textbf{86.987}        & \textbf{1}          & 18                        \\ \hline
			Proposed                & \textbf{0.913}         & \textbf{1}            & 0.724        & 5           & 0.906         & 2           & 0.785         & 6           & 49.15          & 3          & 133.98               & 4          & 21                        \\ \hline \hline
			\cite{xu2016gland}    & 0.858         & 9            & 0.771        & 3           & 0.888         & 4           & 0.815         & 3           & 54.202         & 5          & 129.93         & 3          & 27                        \\ \hline
			CUMedVision2            & 0.912         & 2           & 0.716         & 7           & 0.897         & 3           & 0.781         & 8           & 45.418         & 2          & 160.347        & 10          & 32                        \\ \hline
			ExB1                    & 0.891         & 6           & 0.703         & 8           & 0.882         & 8           & 0.786         & 5           & 57.413         & 10          & 145.575        & 5          & 42                        \\ \hline
			ExB3                    & 0.896         & 3           & 0.719         & 6           & 0.886         & 6           & 0.765         & 9           & 57.350         & 9          & 159.873        & 9          & 42                        \\ \hline
			Freiburg2               & 0.870         & 7           & 0.695         & 9           & 0.876         & 9           & 0.786         & 5           & 57.093         & 7          & 148.463        & 7          & 44                        \\ \hline
			CUMedVision1            & 0.868         & 8           & 0.769         & 4           & 0.867         & 11           & 0.800         & 4           & 74.596         & 11         & 153.646        & 8          & 46                        \\ \hline
			ExB2                    & 0.892         & 5           & 0.686         & 10           & 0.884         & 7           & 0.754         & 10           & 54.785         & 6          & 187.442        & 12         & 50                        \\ \hline
			Freiburg1               & 0.834         & 10           & 0.605         & 11           & 0.875         & 10           & 0.783         & 7           & 57.194         & 8          & 146.607        & 6          & 52                        \\ \hline
			CVML                    & 0.652         & 12          & 0.541         & 12          & 0.644         & 14          & 0.654         & 11         & 155.433        & 14         & 176.244        & 11          & 74                        \\ \hline
			LIB                     & 0.777         & 11          & 0.306         & 14          & 0.781         & 12          & 0.617         & 12         & 112.706        & 13         & 190.447        & 13         & 75                        \\ \hline
			vision4GlaS             & 0.635         & 13         & 0.527          & 13          & 0.737         & 13          & 0.610         & 13         & 107.491        & 12         & 210.105        & 14         & 78                        \\ \hline
		\end{tabular}
	\end{sidewaystable}
	
	\section{Conclusions}
	\label{sec:conclusions}
	In multi-channel fluorescence microscopy, cell segmentation can help to build molecular profiles of individual cells. However, images captured using fluorescence microscopy contain very weak and variable intensities that make it difficult to segment cells in these types of images. The variable size of the cells makes it even more challenging for image processing algorithms to perform cell segmentation.  In tumour histology slides, the morphology of nuclei and nuclear pleomorphism can help in making a diagnosis and in studying the tumour microenvironment but nuclear segmentation is difficult due to varying shape, size, chromatin structure and clumped nuclei. Similarly morphology of glands can help the pathologist to grade the cancer but it is also very challenging due to texture, size and structure of the glands. All these tasks require sophisticated segmentation algorithms. We have presented a deep learning architecture named Micro-Net that can be used to segment cells/nuclei and glands in fluorescence and H\&E stained images with slight tuning of the input parameters. The proposed architecture allows the network to visualise input and output at multiple resolutions. The extra convolutional layers bypass the max-pooling layer, thus allowing the network to better train its parameters for weak features in addition to the strongly observed feature sets. This has been demonstrated by the robustness of algorithm to varying level of noise in fluorescence image data. Intermediate connections between the layers allow context and localization to be retained. The qualitative and quantitative results show that the Micro-Net architecture outperforms recently published deep learning approaches. We will make the fluorescence image data set publicly available subject to publication of this manuscript. The other two data sets we used in this paper are already publicly available. We showed that the proposed algorithm produces competitive results compared to the state-of-the-art. The results produced by the algorithm can be extended to build molecular profile in multiplexed fluorescence images, grade cancer or study tumour microenvironment. 
	
	\section{Acknowledgements}
	\label{sec:Ack}
	We are grateful to the BBSRC UK for supporting this study through project grant BB/K018868/1.

	
	\bibliography{ref}

\begin{thebibliography}{39}
\expandafter\ifx\csname natexlab\endcsname\relax\def\natexlab#1{#1}\fi
\providecommand{\url}[1]{\texttt{#1}}
\providecommand{\href}[2]{#2}
\providecommand{\path}[1]{#1}
\providecommand{\DOIprefix}{doi:}
\providecommand{\ArXivprefix}{arXiv:}
\providecommand{\URLprefix}{URL: }
\providecommand{\Pubmedprefix}{pmid:}
\providecommand{\doi}[1]{\href{http://dx.doi.org/#1}{\path{#1}}}
\providecommand{\Pubmed}[1]{\href{pmid:#1}{\path{#1}}}
\providecommand{\bibinfo}[2]{#2}
\ifx\xfnm\undefined \def\xfnm[#1]{\unskip,\space#1}\fi
\bibitem[{CPM()}]{CPMwebsite}
\bibinfo{title}{Computational precision medicine nuclei segmentation challenge
  website}.
\newblock
  \bibinfo{howpublished}{\url{http://miccai.cloudapp.net/competitions/57}}.
\bibitem[{Abadi et~al.(2015)Abadi, Agarwal, Barham, Brevdo, Chen, Citro,
  Corrado, Davis, Dean, Devin, Ghemawat, Goodfellow, Harp, Irving, Isard, Jia,
  Jozefowicz, Kaiser, Kudlur, Levenberg, Man\'{e}, Monga, Moore, Murray, Olah,
  Schuster, Shlens, Steiner, Sutskever, Talwar, Tucker, Vanhoucke, Vasudevan,
  Vi\'{e}gas, Vinyals, Warden, Wattenberg, Wicke, Yu and
  Zheng}]{tensorflow2015-whitepaper}
\bibinfo{author}{Abadi\xfnm[ M.]}, \bibinfo{author}{Agarwal\xfnm[ A.]},
  \bibinfo{author}{Barham\xfnm[ P.]}, \bibinfo{author}{Brevdo\xfnm[ E.]},
  \bibinfo{author}{Chen\xfnm[ Z.]}, \bibinfo{author}{Citro\xfnm[ C.]},
  \bibinfo{author}{Corrado\xfnm[ G.S.]}, \bibinfo{author}{Davis\xfnm[ A.]},
  \bibinfo{author}{Dean\xfnm[ J.]}, \bibinfo{author}{Devin\xfnm[ M.]},
  \bibinfo{author}{Ghemawat\xfnm[ S.]}, \bibinfo{author}{Goodfellow\xfnm[ I.]},
  \bibinfo{author}{Harp\xfnm[ A.]}, \bibinfo{author}{Irving\xfnm[ G.]},
  \bibinfo{author}{Isard\xfnm[ M.]}, \bibinfo{author}{Jia\xfnm[ Y.]},
  \bibinfo{author}{Jozefowicz\xfnm[ R.]}, \bibinfo{author}{Kaiser\xfnm[ L.]},
  \bibinfo{author}{Kudlur\xfnm[ M.]}, \bibinfo{author}{Levenberg\xfnm[ J.]},
  \bibinfo{author}{Man\'{e}\xfnm[ D.]}, \bibinfo{author}{Monga\xfnm[ R.]},
  \bibinfo{author}{Moore\xfnm[ S.]}, \bibinfo{author}{Murray\xfnm[ D.]},
  \bibinfo{author}{Olah\xfnm[ C.]}, \bibinfo{author}{Schuster\xfnm[ M.]},
  \bibinfo{author}{Shlens\xfnm[ J.]}, \bibinfo{author}{Steiner\xfnm[ B.]},
  \bibinfo{author}{Sutskever\xfnm[ I.]}, \bibinfo{author}{Talwar\xfnm[ K.]},
  \bibinfo{author}{Tucker\xfnm[ P.]}, \bibinfo{author}{Vanhoucke\xfnm[ V.]},
  \bibinfo{author}{Vasudevan\xfnm[ V.]}, \bibinfo{author}{Vi\'{e}gas\xfnm[
  F.]}, \bibinfo{author}{Vinyals\xfnm[ O.]}, \bibinfo{author}{Warden\xfnm[
  P.]}, \bibinfo{author}{Wattenberg\xfnm[ M.]}, \bibinfo{author}{Wicke\xfnm[
  M.]}, \bibinfo{author}{Yu\xfnm[ Y.]}, \bibinfo{author}{Zheng\xfnm[ X.]}.
\newblock \bibinfo{title}{{TensorFlow}: Large-scale machine learning on
  heterogeneous systems}.
\newblock \bibinfo{year}{2015}.
\newblock \URLprefix \url{https://www.tensorflow.org/}; \bibinfo{note}{software
  available from tensorflow.org}.
\bibitem[{Awan et~al.(2017)Awan, Sirinukunwattana, Epstein, Jefferyes, Qidwai,
  Aftab, Mujeeb, Snead and Rajpoot}]{awan2017glandular}
\bibinfo{author}{Awan\xfnm[ R.]}, \bibinfo{author}{Sirinukunwattana\xfnm[ K.]},
  \bibinfo{author}{Epstein\xfnm[ D.]}, \bibinfo{author}{Jefferyes\xfnm[ S.]},
  \bibinfo{author}{Qidwai\xfnm[ U.]}, \bibinfo{author}{Aftab\xfnm[ Z.]},
  \bibinfo{author}{Mujeeb\xfnm[ I.]}, \bibinfo{author}{Snead\xfnm[ D.]},
  \bibinfo{author}{Rajpoot\xfnm[ N.]}.
\newblock \bibinfo{title}{Glandular morphometrics for objective grading of
  colorectal adenocarcinoma histology images}.
\newblock \bibinfo{journal}{Scientific reports}
  \bibinfo{year}{2017};\bibinfo{volume}{7}(\bibinfo{number}{1}):\bibinfo{pages}{16852}.
\newblock \DOIprefix\doi{10.1038/s41598-017-16516-w}.
\bibitem[{Bergeest and Rohr(2012)}]{bergeest2012efficient}
\bibinfo{author}{Bergeest\xfnm[ J.]}, \bibinfo{author}{Rohr\xfnm[ K.]}.
\newblock \bibinfo{title}{Efficient globally optimal segmentation of cells in
  fluorescence microscopy images using level sets and convex energy
  functionals}.
\newblock \bibinfo{journal}{Medical image analysis}
  \bibinfo{year}{2012};\bibinfo{volume}{16}(\bibinfo{number}{7}):\bibinfo{pages}{1436--1444}.
\newblock \DOIprefix\doi{10.1016/j.media.2012.05.012}.
\bibitem[{Carpenter et~al.(2006)Carpenter, Jones, Lamprecht, Clarke, Kang,
  Friman, Guertin, Chang, Lindquist, Moffat et~al.}]{cellprofiler2006}
\bibinfo{author}{Carpenter\xfnm[ A.E.]}, \bibinfo{author}{Jones\xfnm[ T.R.]},
  \bibinfo{author}{Lamprecht\xfnm[ M.R.]}, \bibinfo{author}{Clarke\xfnm[ C.]},
  \bibinfo{author}{Kang\xfnm[ I.H.]}, \bibinfo{author}{Friman\xfnm[ O.]},
  \bibinfo{author}{Guertin\xfnm[ D.A.]}, \bibinfo{author}{Chang\xfnm[ J.H.]},
  \bibinfo{author}{Lindquist\xfnm[ R.A.]}, \bibinfo{author}{Moffat\xfnm[ J.]},
  et~al.
\newblock \bibinfo{title}{Cellprofiler: image analysis software for identifying
  and quantifying cell phenotypes}.
\newblock \bibinfo{journal}{Genome biology}
  \bibinfo{year}{2006};\bibinfo{volume}{7}(\bibinfo{number}{10}):\bibinfo{pages}{R100}.
\newblock \DOIprefix\doi{10.1186/gb-2006-7-10-r100}.
\bibitem[{Chen et~al.(2017)Chen, Qi, Yu, Dou, Qin and Heng}]{Chen2017dcan}
\bibinfo{author}{Chen\xfnm[ H.]}, \bibinfo{author}{Qi\xfnm[ X.]},
  \bibinfo{author}{Yu\xfnm[ L.]}, \bibinfo{author}{Dou\xfnm[ Q.]},
  \bibinfo{author}{Qin\xfnm[ J.]}, \bibinfo{author}{Heng\xfnm[ P.A.]}.
\newblock \bibinfo{title}{Dcan: Deep contour-aware networks for object instance
  segmentation from histology images}.
\newblock \bibinfo{journal}{Medical Image Analysis}
  \bibinfo{year}{2017};\bibinfo{volume}{36}(\bibinfo{number}{Supplement
  C}):\bibinfo{pages}{135 -- 146}.
\newblock \DOIprefix\doi{10.1016/j.media.2016.11.004}.
\bibitem[{Cohen et~al.(2015)Cohen, Rivlin, Shimshoni and
  Sabo}]{cohen2015memory}
\bibinfo{author}{Cohen\xfnm[ A.]}, \bibinfo{author}{Rivlin\xfnm[ E.]},
  \bibinfo{author}{Shimshoni\xfnm[ I.]}, \bibinfo{author}{Sabo\xfnm[ E.]}.
\newblock \bibinfo{title}{Memory based active contour algorithm using
  pixel-level classified images for colon crypt segmentation}.
\newblock \bibinfo{journal}{Computerized Medical Imaging and Graphics}
  \bibinfo{year}{2015};\bibinfo{volume}{43}:\bibinfo{pages}{150--164}.
\newblock \DOIprefix\doi{10.1016/j.compmedimag.2014.12.006}.
\bibitem[{Dimopoulos et~al.(2014)Dimopoulos, Mayer, Rudolf and
  Stelling}]{Dimopoulos15092014}
\bibinfo{author}{Dimopoulos\xfnm[ S.]}, \bibinfo{author}{Mayer\xfnm[ C.E.]},
  \bibinfo{author}{Rudolf\xfnm[ F.]}, \bibinfo{author}{Stelling\xfnm[ J.]}.
\newblock \bibinfo{title}{Accurate cell segmentation in microscopy images using
  membrane patterns}.
\newblock \bibinfo{journal}{Bioinformatics}
  \bibinfo{year}{2014};\bibinfo{volume}{30}(\bibinfo{number}{18}):\bibinfo{pages}{2644--2651}.
\newblock \DOIprefix\doi{10.1093/bioinformatics/btu302}.
\bibitem[{Farjam et~al.(2007)Farjam, Soltanian-Zadeh, Jafari-Khouzani and
  Zoroofi}]{Farjam2007}
\bibinfo{author}{Farjam\xfnm[ R.]}, \bibinfo{author}{Soltanian-Zadeh\xfnm[
  H.]}, \bibinfo{author}{Jafari-Khouzani\xfnm[ K.]},
  \bibinfo{author}{Zoroofi\xfnm[ R.A.]}.
\newblock \bibinfo{title}{An image analysis approach for automatic malignancy
  determination of prostate pathological images}.
\newblock \bibinfo{journal}{Cytometry Part B: Clinical Cytometry}
  \bibinfo{year}{2007};\bibinfo{volume}{72B}(\bibinfo{number}{4}):\bibinfo{pages}{227--240}.
\newblock \DOIprefix\doi{10.1002/cyto.b.20162}.
\bibitem[{Graham and Rajpoot(2018)}]{Graham2018}
\bibinfo{author}{Graham\xfnm[ S.]}, \bibinfo{author}{Rajpoot\xfnm[ N.M.]}.
\newblock \bibinfo{title}{Sams-net: Stain-aware multi-scale network for
  instance-based nuclei segmentation in histology images}.
\newblock In: \bibinfo{booktitle}{2018 IEEE 15th International Symposium on
  Biomedical Imaging (ISBI 2018)}. \bibinfo{year}{2018}. p.
  \bibinfo{pages}{590--594}.
\newblock \DOIprefix\doi{10.1109/ISBI.2018.8363645}.
\bibitem[{Gunduz-Demir et~al.(2010)Gunduz-Demir, Kandemir, Tosun and
  Sokmensuer}]{gunduz2010automatic}
\bibinfo{author}{Gunduz-Demir\xfnm[ C.]}, \bibinfo{author}{Kandemir\xfnm[ M.]},
  \bibinfo{author}{Tosun\xfnm[ A.B.]}, \bibinfo{author}{Sokmensuer\xfnm[ C.]}.
\newblock \bibinfo{title}{Automatic segmentation of colon glands using
  object-graphs}.
\newblock \bibinfo{journal}{Medical image analysis}
  \bibinfo{year}{2010};\bibinfo{volume}{14}(\bibinfo{number}{1}):\bibinfo{pages}{1--12}.
\newblock \DOIprefix\doi{10.1016/j.media.2009.09.001}.
\bibitem[{He et~al.(2016)He, Zhang, Ren and Sun}]{he2016deep}
\bibinfo{author}{He\xfnm[ K.]}, \bibinfo{author}{Zhang\xfnm[ X.]},
  \bibinfo{author}{Ren\xfnm[ S.]}, \bibinfo{author}{Sun\xfnm[ J.]}.
\newblock \bibinfo{title}{Deep residual learning for image recognition}.
\newblock In: \bibinfo{booktitle}{Proceedings of the IEEE conference on
  computer vision and pattern recognition}. \bibinfo{year}{2016}. p.
  \bibinfo{pages}{770--778}.
\newblock \DOIprefix\doi{10.1109/CVPR.2016.90}.
\bibitem[{Kraus et~al.(2016)Kraus, Ba and Frey}]{kraus2016}
\bibinfo{author}{Kraus\xfnm[ O.Z.]}, \bibinfo{author}{Ba\xfnm[ J.L.]},
  \bibinfo{author}{Frey\xfnm[ B.J.]}.
\newblock \bibinfo{title}{Classifying and segmenting microscopy images with
  deep multiple instance learning}.
\newblock \bibinfo{journal}{Bioinformatics}
  \bibinfo{year}{2016};\bibinfo{volume}{32}(\bibinfo{number}{12}):\bibinfo{pages}{i52--i59}.
\newblock \DOIprefix\doi{10.1093/bioinformatics/btw252}.
\bibitem[{Li et~al.(2017)Li, Raza and Rajpoot}]{Li2017multi}
\bibinfo{author}{Li\xfnm[ G.]}, \bibinfo{author}{Raza\xfnm[ S.E.A.]},
  \bibinfo{author}{Rajpoot\xfnm[ N.M.]}.
\newblock \bibinfo{title}{Multi-resolution cell orientation congruence
  descriptors for epithelium segmentation in endometrial histology images}.
\newblock \bibinfo{journal}{Medical Image Analysis}
  \bibinfo{year}{2017};\bibinfo{volume}{37}:\bibinfo{pages}{91 -- 100}.
\newblock \DOIprefix\doi{10.1016/j.media.2017.01.006}.
\bibitem[{Li et~al.(2015)Li, Sanchez, Nagaraj, Khan and Rajpoot}]{Li2015}
\bibinfo{author}{Li\xfnm[ G.]}, \bibinfo{author}{Sanchez\xfnm[ V.]},
  \bibinfo{author}{Nagaraj\xfnm[ P.]}, \bibinfo{author}{Khan\xfnm[ S.]},
  \bibinfo{author}{Rajpoot\xfnm[ N.]}.
\newblock \bibinfo{title}{A novel multitarget tracking algorithm for myosin vi
  protein molecules on actin filaments in tirfm sequences}.
\newblock \bibinfo{journal}{Journal of Microscopy}
  \bibinfo{year}{2015};\bibinfo{volume}{260}(\bibinfo{number}{3}):\bibinfo{pages}{312--325}.
\newblock \DOIprefix\doi{10.1111/jmi.12299}.
\bibitem[{Manivannan et~al.(2018)Manivannan, Li, Zhang, Trucco and
  McKenna}]{Manivannan2017}
\bibinfo{author}{Manivannan\xfnm[ S.]}, \bibinfo{author}{Li\xfnm[ W.]},
  \bibinfo{author}{Zhang\xfnm[ J.]}, \bibinfo{author}{Trucco\xfnm[ E.]},
  \bibinfo{author}{McKenna\xfnm[ S.J.]}.
\newblock \bibinfo{title}{Structure prediction for gland segmentation with
  hand-crafted and deep convolutional features}.
\newblock \bibinfo{journal}{IEEE Transactions on Medical Imaging}
  \bibinfo{year}{2018};\bibinfo{volume}{37}(\bibinfo{number}{1}):\bibinfo{pages}{210--221}.
\newblock \DOIprefix\doi{10.1109/TMI.2017.2750210}.
\bibitem[{Meijering(2012)}]{Meijering2012}
\bibinfo{author}{Meijering\xfnm[ E.]}.
\newblock \bibinfo{title}{{Cell segmentation: 50 years down the road}}.
\newblock \bibinfo{journal}{Signal Processing Magazine, IEEE}
  \bibinfo{year}{2012};\bibinfo{volume}{29}(\bibinfo{number}{5}):\bibinfo{pages}{140--145}.
\newblock \DOIprefix\doi{10.1109/MSP.2012.2204190}.
\bibitem[{Naik et~al.(2008)Naik, Doyle, Agner, Madabhushi, Feldman and
  Tomaszewski}]{naik2008automated}
\bibinfo{author}{Naik\xfnm[ S.]}, \bibinfo{author}{Doyle\xfnm[ S.]},
  \bibinfo{author}{Agner\xfnm[ S.]}, \bibinfo{author}{Madabhushi\xfnm[ A.]},
  \bibinfo{author}{Feldman\xfnm[ M.]}, \bibinfo{author}{Tomaszewski\xfnm[ J.]}.
\newblock \bibinfo{title}{Automated gland and nuclei segmentation for grading
  of prostate and breast cancer histopathology}.
\newblock In: \bibinfo{booktitle}{Biomedical Imaging: From Nano to Macro, 2008.
  ISBI 2008. 5th IEEE International Symposium on}.
  \bibinfo{organization}{IEEE}; \bibinfo{year}{2008}. p.
  \bibinfo{pages}{284--287}.
\newblock \DOIprefix\doi{10.1109/ISBI.2008.4540988}.
\bibitem[{Nguyen et~al.(2012)Nguyen, Sarkar and Jain}]{nguyen2012structure}
\bibinfo{author}{Nguyen\xfnm[ K.]}, \bibinfo{author}{Sarkar\xfnm[ A.]},
  \bibinfo{author}{Jain\xfnm[ A.K.]}.
\newblock \bibinfo{title}{Structure and context in prostatic gland segmentation
  and classification}.
\newblock In: \bibinfo{booktitle}{Medical Image Computing and Computer-Assisted
  Intervention--MICCAI 2012}. \bibinfo{publisher}{Springer};
  \bibinfo{year}{2012}. p. \bibinfo{pages}{115--123}.
\newblock \DOIprefix\doi{10.1007/978-3-642-33415-3_15}.
\bibitem[{Nosrati and Hamarneh(2014)}]{nosrati2014local}
\bibinfo{author}{Nosrati\xfnm[ M.S.]}, \bibinfo{author}{Hamarneh\xfnm[ G.]}.
\newblock \bibinfo{title}{Local optimization based segmentation of
  spatially-recurring, multi-region objects with part configuration
  constraints}.
\newblock \bibinfo{journal}{IEEE transactions on medical imaging}
  \bibinfo{year}{2014};\bibinfo{volume}{33}(\bibinfo{number}{9}):\bibinfo{pages}{1845--1859}.
\newblock \DOIprefix\doi{10.1109/TMI.2014.2323074}.
\bibitem[{Qaiser et~al.(2017)Qaiser, Tsang, Epstein and
  Rajpoot}]{qaiser2017tumor}
\bibinfo{author}{Qaiser\xfnm[ T.]}, \bibinfo{author}{Tsang\xfnm[ Y.W.]},
  \bibinfo{author}{Epstein\xfnm[ D.]}, \bibinfo{author}{Rajpoot\xfnm[ N.]}.
\newblock \bibinfo{title}{Tumor segmentation in whole slide images using
  persistent homology and deep convolutional features}.
\newblock In: \bibinfo{booktitle}{Annual Conference on Medical Image
  Understanding and Analysis}. \bibinfo{organization}{Springer};
  \bibinfo{year}{2017}. p. \bibinfo{pages}{320--329}.
\newblock \DOIprefix\doi{10.1007/978-3-319-60964-5_28}.
\bibitem[{Raza et~al.(2017{\natexlab{a}})Raza, Cheung, Epstein, Pelengaris,
  Khan and Rajpoot}]{RazaISBI2017}
\bibinfo{author}{Raza\xfnm[ S.E.A.]}, \bibinfo{author}{Cheung\xfnm[ L.]},
  \bibinfo{author}{Epstein\xfnm[ D.]}, \bibinfo{author}{Pelengaris\xfnm[ S.]},
  \bibinfo{author}{Khan\xfnm[ M.]}, \bibinfo{author}{Rajpoot\xfnm[ N.M.]}.
\newblock \bibinfo{title}{Mimo-net: A multi-input multi-output convolutional
  neural network for cell segmentation in fluorescence microscopy images}.
\newblock In: \bibinfo{booktitle}{2017 IEEE 14th International Symposium on
  Biomedical Imaging (ISBI 2017)}. \bibinfo{year}{2017}{\natexlab{a}}. p.
  \bibinfo{pages}{337--340}.
\newblock \DOIprefix\doi{10.1109/ISBI.2017.7950532}.
\bibitem[{Raza et~al.(2017{\natexlab{b}})Raza, Cheung, Epstein, Pelengaris,
  Khan and Rajpoot}]{RazaMIUA2017}
\bibinfo{author}{Raza\xfnm[ S.E.A.]}, \bibinfo{author}{Cheung\xfnm[ L.]},
  \bibinfo{author}{Epstein\xfnm[ D.]}, \bibinfo{author}{Pelengaris\xfnm[ S.]},
  \bibinfo{author}{Khan\xfnm[ M.]}, \bibinfo{author}{Rajpoot\xfnm[ N.M.]}.
\newblock \bibinfo{title}{MIMONet: Gland Segmentation Using
  Multi-Input-Multi-Output Convolutional Neural Network};
  \bibinfo{address}{Cham}: \bibinfo{publisher}{Springer International
  Publishing}.
\newblock p. \bibinfo{pages}{698--706}.
\newblock \DOIprefix\doi{10.1007/978-3-319-60964-5_61}.
\bibitem[{Raza et~al.(2012)Raza, Humayun, Abouna, Nattkemper, Epstein, Khan,
  Rajpoot et~al.}]{Raza2012}
\bibinfo{author}{Raza\xfnm[ S.E.A.]}, \bibinfo{author}{Humayun\xfnm[ A.]},
  \bibinfo{author}{Abouna\xfnm[ S.]}, \bibinfo{author}{Nattkemper\xfnm[ T.W.]},
  \bibinfo{author}{Epstein\xfnm[ D.B.]}, \bibinfo{author}{Khan\xfnm[ M.]},
  \bibinfo{author}{Rajpoot\xfnm[ N.M.]}, et~al.
\newblock \bibinfo{title}{{RAMTaB: robust alignment of multi-tag bioimages.}}
\newblock \bibinfo{journal}{PLoS ONE}
  \bibinfo{year}{2012};\bibinfo{volume}{7}(\bibinfo{number}{2}):\bibinfo{pages}{e30894}.
\newblock \DOIprefix\doi{10.1371/journal.pone.0030894}.
\bibitem[{Raza et~al.(2016)Raza, Langenk{\"a}mper, Sirinukunwattana, Epstein,
  Nattkemper and Rajpoot}]{Raza2016}
\bibinfo{author}{Raza\xfnm[ S.E.A.]}, \bibinfo{author}{Langenk{\"a}mper\xfnm[
  D.]}, \bibinfo{author}{Sirinukunwattana\xfnm[ K.]},
  \bibinfo{author}{Epstein\xfnm[ D.]}, \bibinfo{author}{Nattkemper\xfnm[
  T.W.]}, \bibinfo{author}{Rajpoot\xfnm[ N.M.]}.
\newblock \bibinfo{title}{{Robust normalization protocols for multiplexed
  fluorescence bioimage analysis}}.
\newblock \bibinfo{journal}{BioData Mining}
  \bibinfo{year}{2016};\bibinfo{volume}{9}(\bibinfo{number}{1}):\bibinfo{pages}{11}.
\newblock \DOIprefix\doi{10.1186/s13040-016-0088-2}.
\bibitem[{Reinhard et~al.(2001)Reinhard, Adhikhmin, Gooch and
  Shirley}]{reinhard2001color}
\bibinfo{author}{Reinhard\xfnm[ E.]}, \bibinfo{author}{Adhikhmin\xfnm[ M.]},
  \bibinfo{author}{Gooch\xfnm[ B.]}, \bibinfo{author}{Shirley\xfnm[ P.]}.
\newblock \bibinfo{title}{Color transfer between images}.
\newblock \bibinfo{journal}{IEEE Computer graphics and applications}
  \bibinfo{year}{2001};\bibinfo{volume}{21}(\bibinfo{number}{5}):\bibinfo{pages}{34--41}.
\bibitem[{Ronneberger et~al.(2015)Ronneberger, Fischer and
  Brox}]{Ronneberger2015}
\bibinfo{author}{Ronneberger\xfnm[ O.]}, \bibinfo{author}{Fischer\xfnm[ P.]},
  \bibinfo{author}{Brox\xfnm[ T.]}.
\newblock \bibinfo{title}{U-net: Convolutional networks for biomedical image
  segmentation}.
\newblock In: \bibinfo{booktitle}{International Conference on Medical Image
  Computing and Computer-Assisted Intervention}. \bibinfo{year}{2015}. p.
  \bibinfo{pages}{234--241}.
\newblock \DOIprefix\doi{10.1007/978-3-319-24574-4_28}.
\bibitem[{Sadanandan et~al.(2017)Sadanandan, Ranefall, Le~Guyader and
  W{\"a}hlby}]{sadanandan2017}
\bibinfo{author}{Sadanandan\xfnm[ S.K.]}, \bibinfo{author}{Ranefall\xfnm[ P.]},
  \bibinfo{author}{Le~Guyader\xfnm[ S.]}, \bibinfo{author}{W{\"a}hlby\xfnm[
  C.]}.
\newblock \bibinfo{title}{Automated training of deep convolutional neural
  networks for cell segmentation}.
\newblock \bibinfo{journal}{Scientific reports}
  \bibinfo{year}{2017};\bibinfo{volume}{7}(\bibinfo{number}{1}):\bibinfo{pages}{7860}.
\newblock \DOIprefix\doi{10.1038/s41598-017-07599-6}.
\bibitem[{Schubert et~al.(2006)Schubert, Bonnekoh, Pommer, Philipsen,
  B{\"o}ckelmann, Malykh, Gollnick, Friedenberger, Bode and
  Dress}]{schubert2006}
\bibinfo{author}{Schubert\xfnm[ W.]}, \bibinfo{author}{Bonnekoh\xfnm[ B.]},
  \bibinfo{author}{Pommer\xfnm[ A.J.]}, \bibinfo{author}{Philipsen\xfnm[ L.]},
  \bibinfo{author}{B{\"o}ckelmann\xfnm[ R.]}, \bibinfo{author}{Malykh\xfnm[
  Y.]}, \bibinfo{author}{Gollnick\xfnm[ H.]},
  \bibinfo{author}{Friedenberger\xfnm[ M.]}, \bibinfo{author}{Bode\xfnm[ M.]},
  \bibinfo{author}{Dress\xfnm[ A.W.]}.
\newblock \bibinfo{title}{Analyzing proteome topology and function by automated
  multidimensional fluorescence microscopy}.
\newblock \bibinfo{journal}{Nature biotechnology}
  \bibinfo{year}{2006};\bibinfo{volume}{24}(\bibinfo{number}{10}):\bibinfo{pages}{1270}.
\newblock \DOIprefix\doi{10.1038/nbt1250}.
\bibitem[{Shelhamer et~al.(2017)Shelhamer, Long and Darrell}]{FCN2016}
\bibinfo{author}{Shelhamer\xfnm[ E.]}, \bibinfo{author}{Long\xfnm[ J.]},
  \bibinfo{author}{Darrell\xfnm[ T.]}.
\newblock \bibinfo{title}{Fully convolutional networks for semantic
  segmentation}.
\newblock \bibinfo{journal}{IEEE Transactions on Pattern Analysis and Machine
  Intelligence}
  \bibinfo{year}{2017};\bibinfo{volume}{39}(\bibinfo{number}{4}):\bibinfo{pages}{640--651}.
\newblock \DOIprefix\doi{10.1109/TPAMI.2016.2572683}.
\bibitem[{Sirinukunwattana et~al.(2017)Sirinukunwattana, Pluim, Chen, Qi, Heng,
  Guo, Wang, Matuszewski, Bruni, Sanchez et~al.}]{Sirinukunwattana2016GLaS}
\bibinfo{author}{Sirinukunwattana\xfnm[ K.]}, \bibinfo{author}{Pluim\xfnm[
  J.P.]}, \bibinfo{author}{Chen\xfnm[ H.]}, \bibinfo{author}{Qi\xfnm[ X.]},
  \bibinfo{author}{Heng\xfnm[ P.A.]}, \bibinfo{author}{Guo\xfnm[ Y.B.]},
  \bibinfo{author}{Wang\xfnm[ L.Y.]}, \bibinfo{author}{Matuszewski\xfnm[
  B.J.]}, \bibinfo{author}{Bruni\xfnm[ E.]}, \bibinfo{author}{Sanchez\xfnm[
  U.]}, et~al.
\newblock \bibinfo{title}{Gland segmentation in colon histology images: The
  glas challenge contest}.
\newblock \bibinfo{journal}{Medical image analysis}
  \bibinfo{year}{2017};\bibinfo{volume}{35}:\bibinfo{pages}{489--502}.
\newblock \DOIprefix\doi{10.1016/j.media.2016.08.008}.
\bibitem[{Sirinukunwattana et~al.(2015)Sirinukunwattana, Snead and
  Rajpoot}]{sirinukunwattana2015stochastic}
\bibinfo{author}{Sirinukunwattana\xfnm[ K.]}, \bibinfo{author}{Snead\xfnm[
  D.R.]}, \bibinfo{author}{Rajpoot\xfnm[ N.M.]}.
\newblock \bibinfo{title}{A stochastic polygons model for glandular structures
  in colon histology images}.
\newblock \bibinfo{journal}{IEEE transactions on medical imaging}
  \bibinfo{year}{2015};\bibinfo{volume}{34}(\bibinfo{number}{11}):\bibinfo{pages}{2366--2378}.
\newblock \DOIprefix\doi{10.1109/TMI.2015.2433900}.
\bibitem[{Song et~al.(2017)Song, Tan, Jiang, Cheng, Ni, Chen, Lei and
  Wang}]{Song2016}
\bibinfo{author}{Song\xfnm[ Y.]}, \bibinfo{author}{Tan\xfnm[ E.L.]},
  \bibinfo{author}{Jiang\xfnm[ X.]}, \bibinfo{author}{Cheng\xfnm[ J.Z.]},
  \bibinfo{author}{Ni\xfnm[ D.]}, \bibinfo{author}{Chen\xfnm[ S.]},
  \bibinfo{author}{Lei\xfnm[ B.]}, \bibinfo{author}{Wang\xfnm[ T.]}.
\newblock \bibinfo{title}{Accurate cervical cell segmentation from overlapping
  clumps in pap smear images}.
\newblock \bibinfo{journal}{IEEE Transactions on Medical Imaging}
  \bibinfo{year}{2017};\bibinfo{volume}{36}(\bibinfo{number}{1}):\bibinfo{pages}{288--300}.
\newblock \DOIprefix\doi{10.1109/TMI.2016.2606380}.
\bibitem[{Veta et~al.(2013)Veta, Van~Diest, Kornegoor, Huisman, Viergever and
  Pluim}]{Veta2013}
\bibinfo{author}{Veta\xfnm[ M.]}, \bibinfo{author}{Van~Diest\xfnm[ P.J.]},
  \bibinfo{author}{Kornegoor\xfnm[ R.]}, \bibinfo{author}{Huisman\xfnm[ A.]},
  \bibinfo{author}{Viergever\xfnm[ M.A.]}, \bibinfo{author}{Pluim\xfnm[ J.P.]}.
\newblock \bibinfo{title}{{Automatic Nuclei Segmentation in H{\&}E Stained
  Breast Cancer Histopathology Images}}.
\newblock \bibinfo{journal}{PLoS ONE}
  \bibinfo{year}{2013};\bibinfo{volume}{8}(\bibinfo{number}{7}):\bibinfo{pages}{e70221}.
\newblock \DOIprefix\doi{10.1371/journal.pone.0070221}.
\bibitem[{Wu et~al.(2005)Wu, Xu, Harpaz, Burstein and Gil}]{Wu2005}
\bibinfo{author}{Wu\xfnm[ H.S.]}, \bibinfo{author}{Xu\xfnm[ R.]},
  \bibinfo{author}{Harpaz\xfnm[ N.]}, \bibinfo{author}{Burstein\xfnm[ D.]},
  \bibinfo{author}{Gil\xfnm[ J.]}.
\newblock \bibinfo{title}{Segmentation of intestinal gland images with
  iterative region growing}.
\newblock \bibinfo{journal}{Journal of Microscopy}
  \bibinfo{year}{2005};\bibinfo{volume}{220}(\bibinfo{number}{3}):\bibinfo{pages}{190--204}.
\newblock \DOIprefix\doi{10.1111/j.1365-2818.2005.01531.x}.
\bibitem[{Xu et~al.(2016)Xu, Li, Liu, Wang, Lai, Eric and Chang}]{xu2016gland}
\bibinfo{author}{Xu\xfnm[ Y.]}, \bibinfo{author}{Li\xfnm[ Y.]},
  \bibinfo{author}{Liu\xfnm[ M.]}, \bibinfo{author}{Wang\xfnm[ Y.]},
  \bibinfo{author}{Lai\xfnm[ M.]}, \bibinfo{author}{Eric\xfnm[ I.]},
  \bibinfo{author}{Chang\xfnm[ C.]}.
\newblock \bibinfo{title}{Gland instance segmentation by deep multichannel side
  supervision}.
\newblock In: \bibinfo{booktitle}{International Conference on Medical Image
  Computing and Computer-Assisted Intervention}.
  \bibinfo{organization}{Springer}; \bibinfo{year}{2016}. p.
  \bibinfo{pages}{496--504}.
\newblock \DOIprefix\doi{10.1007/978-3-319-46723-8_57}.
\bibitem[{Xu et~al.(2017)Xu, Li, Wang, Liu, Fan, Lai and Chang}]{Xu2017}
\bibinfo{author}{Xu\xfnm[ Y.]}, \bibinfo{author}{Li\xfnm[ Y.]},
  \bibinfo{author}{Wang\xfnm[ Y.]}, \bibinfo{author}{Liu\xfnm[ M.]},
  \bibinfo{author}{Fan\xfnm[ Y.]}, \bibinfo{author}{Lai\xfnm[ M.]},
  \bibinfo{author}{Chang\xfnm[ E.I.C.]}.
\newblock \bibinfo{title}{Gland instance segmentation using deep multichannel
  neural networks}.
\newblock \bibinfo{journal}{IEEE Transactions on Biomedical Engineering}
  \bibinfo{year}{2017};\bibinfo{volume}{64}(\bibinfo{number}{12}):\bibinfo{pages}{2901--2912}.
\newblock \DOIprefix\doi{10.1109/TBME.2017.2686418}.
\bibitem[{Yang et~al.(2006)Yang, Li and Zhou}]{yang2006nuclei}
\bibinfo{author}{Yang\xfnm[ X.]}, \bibinfo{author}{Li\xfnm[ H.]},
  \bibinfo{author}{Zhou\xfnm[ X.]}.
\newblock \bibinfo{title}{Nuclei segmentation using marker-controlled
  watershed, tracking using mean-shift, and kalman filter in time-lapse
  microscopy}.
\newblock \bibinfo{journal}{IEEE Transactions on Circuits and Systems I:
  Regular Papers}
  \bibinfo{year}{2006};\bibinfo{volume}{53}(\bibinfo{number}{11}):\bibinfo{pages}{2405--2414}.
\newblock \DOIprefix\doi{10.1109/TCSI.2006.884469}.
\bibitem[{Yuan et~al.(2012)Yuan, Failmezger, Rueda, Ali, Gr{\"a}f, Chin,
  Schwarz, Curtis, Dunning, Bardwell et~al.}]{yuan2012quantitative}
\bibinfo{author}{Yuan\xfnm[ Y.]}, \bibinfo{author}{Failmezger\xfnm[ H.]},
  \bibinfo{author}{Rueda\xfnm[ O.M.]}, \bibinfo{author}{Ali\xfnm[ H.R.]},
  \bibinfo{author}{Gr{\"a}f\xfnm[ S.]}, \bibinfo{author}{Chin\xfnm[ S.F.]},
  \bibinfo{author}{Schwarz\xfnm[ R.F.]}, \bibinfo{author}{Curtis\xfnm[ C.]},
  \bibinfo{author}{Dunning\xfnm[ M.J.]}, \bibinfo{author}{Bardwell\xfnm[ H.]},
  et~al.
\newblock \bibinfo{title}{Quantitative image analysis of cellular heterogeneity
  in breast tumors complements genomic profiling}.
\newblock \bibinfo{journal}{Science translational medicine}
  \bibinfo{year}{2012};\bibinfo{volume}{4}(\bibinfo{number}{157}):\bibinfo{pages}{157ra143--157ra143}.
\newblock \DOIprefix\doi{10.1126/scitranslmed.3004330}.

\end{thebibliography}
	
\end{document}